\DocumentMetadata{testphase=new-or-1}
\documentclass{article} %chktex-file 1 chktex-file 44
\usepackage[hyphens]{url}
\usepackage{hyperref}
\usepackage{styles/iclr2026_conference,styles/multimat}
\usepackage{booktabs,threeparttable,wrapfig2,placeins}
\usepackage[frozencache,cachedir=minted]{minted2}
\usepackage[most,minted]{tcolorbox}
\usepackage{tikz,forest,pgfplots,styles/colorschemes}
\usetikzlibrary{calc,arrows.meta,decorations.pathmorphing}
\pgfplotsset{compat=1.18}
\fvset{backgroundcolorboxoverlap=0pt}

\iclrfinalcopy

\hypersetup{%
  colorlinks=true,
  citecolor=darkblue,
  linkcolor=darkblue,
  urlcolor=darkblue
}

\title{\projectname: Multimodal Program Synthesis for Procedural Materials using Large Multimodal Models}

\author{Jonas Belouadi\\
University of Mannheim, Germany\\
\mail{jonas.belouadi@uni-mannheim.de}\\
\And
Tamy Boubekeur, Adrien Kaiser\\
Adobe Research, France\\
\mails{boubek,akaiser}{adobe.com}
}

\begin{document}

\maketitle

\begin{abstract}
  Material node graphs are programs that generate the 2D channels of procedural
  materials, including geometry such as roughness and displacement maps, and
  reflectance such as albedo and conductivity maps. They are essential in
  computer graphics for representing the appearance of virtual 3D objects
  parametrically and at arbitrary resolution. In particular, their directed
  acyclic graph structure and intermediate states enable a modular,
  interpretable workflow for interactive appearance modeling.
  However, creating such graphs remains challenging and typically requires
  professional training. While recent neural program synthesis approaches
  attempt to simplify this process, they solely represent graphs as
  \emph{textual} programs, failing to capture the inherently visual-spatial
  nature of node graphs that makes them accessible to humans.
  To address this gap, we present \projectname, a \emph{multimodal} program
  synthesis framework that leverages large multimodal models to process both
  visual and textual graph representations for improved generation of
  procedural material graphs. We train our models on a new dataset of
  production-quality procedural materials and combine them with a constrained
  tree search inference algorithm that ensures static correctness while
  efficiently navigating the program space.
  Our experimental results show that our multimodal program synthesis method is
  more efficient in both unconditional and conditional graph synthesis with
  higher visual quality and fidelity than text-only baselines, establishing new
  state-of-the-art performance.
\end{abstract}

\FloatBarrier\section{Introduction}\label{sec:introduction}
\begin{figure*}[t]
    \centering
    \includegraphics[width=\textwidth]{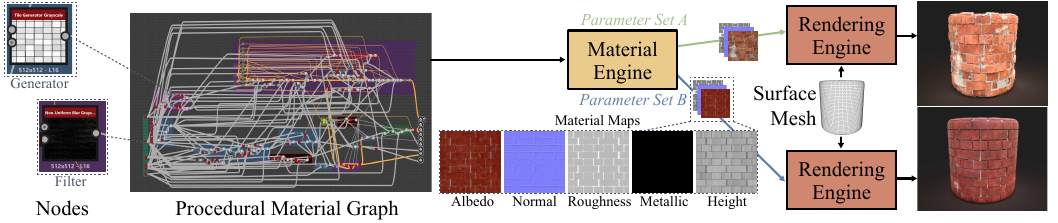}%
    \caption{Procedural materials offer powerful control over the appearance of
      3D objects through a few high-level parameters.  Here, a production-grade
      example (left) with the images obtained using two distinct parameter sets A and B
      (right).
    }%
    \label{fig:proc-mat}
\end{figure*}
Procedural materials have become increasingly important in modern 3D content
creation, offering artists greater control and flexibility in designing surface
appearances for digital assets. Unlike traditional image-based textures, which
are constrained by fixed resolutions and limited editability, procedural
material modeling tools like Adobe Substance
Designer~\citep{SubstanceDesigner2025} or Blender~\citep{Blender2025} leverage
node-based graphs to generate textures programmatically. This enables
resolution-independent execution, high-level parametric control, and
non-destructive editing workflows that have proven valuable in industries such
as game development, film production, and VR/AR
applications~\citep{ebert2002procedural}. More specifically, a procedural
material is defined as a directed graph where nodes represent texture
generators (e.g., noise functions, patterns) or filtering operations (e.g.,
blurs, color adjustments), and edges encode the flow of data between these
operations, ultimately producing the texture maps required by physically-based
rendering (PBR) models~\citep{PBRT3e} (cf.\ \figref{fig:proc-mat}). However, the
complexity of crafting these procedural material graphs presents a substantial
barrier to entry, creating a pressing need for automated and semi-automated
approaches to support material artists at all levels of proficiency.

With recent advances in neural program
synthesis~\citep{huynh2025largelanguagemodelscode}, procedural material
synthesis has become increasingly feasible\@. \matformer pioneered this
direction with a multi-stage transformer-based model for unconditional
generation with Adobe Substance Designer~\citep{guerrero2022matformer}.
Building on this foundation, \citet{hu2023materials} extended the approach to
support conditional synthesis, enabling applications such as inverse
rendering~\citep{patow2003inverse}, i.e., generating procedural materials that
match the appearance of captured or rendered images. More recently, \vlmaterial
demonstrated that large language models~\citep{zhao2025llm} can effectively
perform end-to-end procedural material synthesis~\citep{li2025vlmaterial}.
However, these approaches share a fundamental limitation: they generate node
graphs as text-only programs without access to visual feedback during
synthesis. This contrasts sharply with how human artists work, who create
procedural materials by manipulating node graphs through an arguably more
intuitive visual interface, as illustrated in \figref{fig:proc-mat} (left).
Without visual feedback, models must rely solely on textual representations to
reason about complex spatial relationships and visual outcomes, a task that
becomes increasingly difficult as material complexity grows. To address this
limitation, we propose a novel \emph{multimodal program synthesis} paradigm
based on large multimodal models~\citep{yin2024mllm} that incorporates visual
feedback throughout the generation process, more closely mirroring human
creative workflows. We demonstrate that this approach, to which we refer as
\projectname, outperforms previous state-of-the-art methods~(cf.\
\secref{sec:experiments}). Our key contributions are as follows:
\begin{enumerate}
  \item We introduce \projectname, a novel procedural material synthesis
    approach that incorporates visualizations of intermediate graphs,
    including node states, into its context. This multimodal feedback loop
    improves material quality substantially compared to text-only baselines.
  \item Investigating intermediate states enables real-time validation of each
    generated node. This allows us to develop a tree search algorithm that
    backtracks upon encountering invalid states, enabling more efficient
    inference than prior methods, which often produce invalid graphs.
  \item We implement a transpiler that converts between Adobe Substance
    Designer formats and a compact representation suitable for language
    modeling while supporting the complete feature set. This enables training
    on larger datasets and the generation of more complex materials than
    previous approaches, which examined only limited subsets of Designer's
    capabilities.
\end{enumerate}

\section{Related Work}

\paragraph{Large Language Models for Program Synthesis}
Our work builds upon recent advances in neural program
synthesis~\citep{parisotto2017neurosymbolic,devlin2017robustfill,thakoor2018synthesis,ye2021optimal,ellis2021dreamcoder}.
Traditional program synthesizers require formal specifications and employ
search or logical derivation to produce programs that provably satisfy these
specifications~\citep{alur2013synthesis}. Recently, large language models have
demonstrated impressive capabilities in this
domain~\citep{huynh2025largelanguagemodelscode,li2025toward,lozhkov2024starcoder2stackv2,li2023starcoder,roziere2023code,fried2023incoder,li2022alphacode,chen2021evaluating}.
However, current research predominantly targets high-resource programming
languages such as Python, Java, or
JavaScript~\citep{zan2023lnlcode,huynh2025largelanguagemodelscode}. In
contrast, our work synthesizes \emph{graphics} programs, which pose
unique challenges due to domain-specific requirements and considerable data
scarcity, establishing it as a distinct research area.

\paragraph{Graphics Program Synthesis}
Deep learning approaches have shown strong performance in synthesizing
graphics programs that compile to visual
outputs~\citep{ellis2018drawing,ellis2019repl,ganin2018program}. This progress
has been accelerated by the emergence of large multimodal models, particularly
vision-language models that bridge visual and textual
domains~\citep{alayrac2022flamingo,liu2023visual,belouadi2024detikzify,kulits2024rethinking,li2024is,kapur2025diffusion,lin2025jarvisart,xu2025visualplanningletsthink}.
The field encompasses both controlled experimental settings using
domain-specific
languages~\citep{ellis2018drawing,tian2018learning,sharma2018csgnet,amara2023uml,kulits2024rethinking,kapur2025diffusion,wen2025program}
and practical applications. Notable examples
include systems for generating scientific figures using
\tikzname~\citep{belouadi2024automatikz,belouadi2024detikzify,belouadi2025tikzero,laurencon2024idefics2,laurencon2024idefics3,tong2024cambrian,zhang2024mm15methodsanalysis}
and automating data
visualization~\citep{Mackinlay1986automating,roth1994interactive,luo2021nl2vis,wu2024nl2vis,voigt2024plots}.
However, these approaches generate code designed for text-based editing and
therefore do not face the unique circumstances of node graphs in procedural
material synthesis that our work addresses.

\paragraph{Procedural Material Synthesis}
Procedural material modeling is one of the most challenging domains in
graphics program synthesis. The combination of lengthy, complex material
programs and severe data scarcity creates unique obstacles for learning-based
approaches~\citep{li2025vlmaterial,li2024matformerrl}. Existing methods
primarily focus on inverse procedural material modeling by synthesizing graphs
that reproduce a given target appearance~\citep{hu2023materials} or
unconditional generation to create diverse, novel materials without
specific targets~\citep{guerrero2022matformer}. A related line of work
optimizes parameters of existing material graphs to match image targets by
transpiling them into differentiable programs~\citep{shi2020match,hu2022optim,li2023end}. As
discussed in \secref{sec:introduction}, previous generative approaches are
limited to text-only representations, a limitation we address in this work.

\section{Background on Procedural Materials}\label{sec:background}
As indicated in \secref{sec:introduction}, procedural materials are directed
acyclic graphs $G$, executed by a material engine to produce raster images
representing the physical properties of materials. These so-called material
maps define surface characteristics, e.g., albedo, roughness, or normal
(tangent space orientation), that enable photorealistic rendering when applied
to 3D objects, with their appearance controlled through a small set of
high-level parameters~(cf.\ \figref{fig:proc-mat}). The internal structure of a
material graph $G$ comprises nodes $\{v_1, v_2, \ldots, v_N\}$ connected by
edges that define the flow of image data. Each node $v_i$ functions as either a
generator that creates new image content or a filter that transforms existing
images from upstream nodes. Common node operations include noise generation,
blending, and mathematical transformations, which collectively produce
intermediate image outputs $I=\{i_1, i_2, \ldots, i_N\}$. The behavior of each
node is governed by parameters that may be discrete or continuous scalars or
vectors, providing fine-grained control over the final material appearance.
\begin{figure*}
    \centering
    \includegraphics[width=\textwidth]{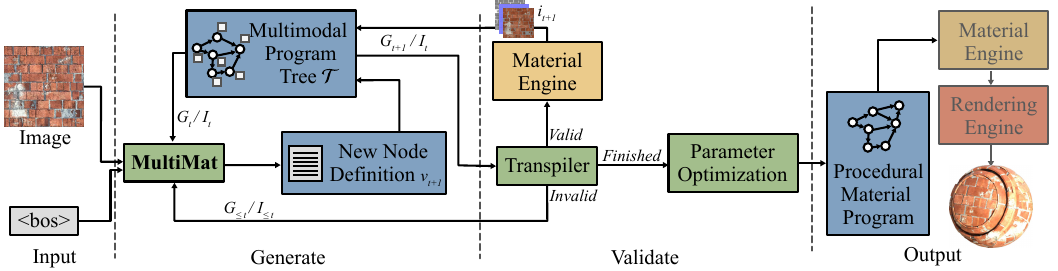}%
    \caption{Architecture overview of \projectname during inference. The system
    constructs a multimodal program tree $\mathcal{T}$ by iteratively
    generating node definitions. At each step $t$, the system derives a graph
    $G_t$ of valid nodes along with corresponding intermediate outputs
    $I_t$ by traversing $\mathcal{T}$, which may contain both
    valid and invalid nodes, to generate the next
    node $v_{t+1}$. When transpilation and execution succeed, the system
    advances with an updated graph $G_{t+1}$ and outputs $I_{t+1}$. If errors
    occur, it reverts to a previous state $(G_{\leq t}, I_{\leq t})$. The
    generation process initiates from either an input image or unconditionally
    using a beginning-of-sequence token (\texttt{<bos>}). Following optional
    parameter optimization (cf.\ \secref{sec:conditional}), the final
    procedural material can be applied to any target geometry for rendering.}%
    \label{fig:pipeline}
\end{figure*}

Professional material authoring tools such as Blender and Adobe Substance
Designer enable artists to construct and modify procedural material graphs
through visual interfaces~(cf.\ \figref{fig:proc-mat}). Users can interactively
add or remove nodes and edges while adjusting node parameters to achieve
desired visual effects. Among these tools, Adobe Substance Designer stands out
for its particularly expressive node graph system, which \projectname
specifically targets. It offers advanced capabilities for creating complex
material appearances through features like function graphs and pixel
processors. Function graphs allow parameters to be controlled through custom
operations on input values, while pixel processors enable users to define
specialized computational graphs that operate on individual pixels using
sequences of atomic mathematical operations. These sophisticated capabilities
make automated procedural material synthesis a particularly challenging problem
in this domain.

\section{The \projectname Model \& Architecture}
\figref{fig:pipeline} illustrates our complete model
pipeline. At its core, \projectname is a vision-language model, trained for
synthesizing procedural material graphs. It accepts images
as input for inverse procedural material synthesis and supports unconditional
generation. Unlike previous approaches, \projectname generates nodes
\emph{topologically}, ensuring each node precedes all nodes it connects to.
This enables an iterative generation process detailed below that can provide
continuous visual feedback to the model, verify the validity of intermediate
outputs, and recover from errors automatically in certain cases.

\subsection{Multimodal Program Synthesis}\label{sec:multimodal-synthesis}
\begin{figure*}
    \centering
    \input{graphics/pipeline-step.tex}%
    \caption{Visualization of the two conditioning approaches used by
    \projectname for generating node definition $v_{t+1}$. In the
    graph-conditioned approach (1), \projectname processes the graph $G_t$ as a
    visual representation similar to human perception. In the mixed-conditioned
    approach (2), \projectname receives $G_t$ as a multimodal program where
    \texttt{<img>} tokens are replaced with their corresponding vision encoder
    representations from $I_t$.}%
    \label{fig:multimat-details}
\end{figure*}
Given a partially generated material graph $G_t = \{v_1, v_2, \ldots, v_t\}$
with nodes $v_i$ at generation step $t$, the topological ordering of nodes
allows for visualizing intermediate node states, similar to visual editing
environments that target humans. This enables an iterative generation loop
where \projectname generates one node definition---including node parameters
and connections to previous nodes---at a time that is processed accordingly
before the generation continues. After generating node $v_{t+1}$ in an
intermediate text format (cf.\ \secref{sec:dataset}), we combine it with the
existing node definitions $\{v_1, \ldots, v_t\}$ and feed them to a transpiler,
which compiles the intermediate representations back to a format the material
engine understands. We then use the material engine to visualize the state of
node $v_{t+1}$. Upon successful transpilation and execution, $v_{t+1}$ is
appended to the graph $G_t$, resulting in $G_{t+1}$. This updated state,
including the visualized intermediate outputs $I_{t+1}$, is fed back to the
model to generate the subsequent node $v_{t+2}$ (cf.\ \figref{fig:pipeline}).
If execution or transpilation fails, we discard the current $v_{t+1}$ and
resample, or backtrack further in case of repeated errors (cf.\
\secref{sec:tree-search}). We explore two complementary approaches for
representing $G_t$ and $I_t$ as \emph{multimodal programs} to the model, as
visualized in \figref{fig:multimat-details}:
\begin{description}
  \item[Mixed Conditioning] Starting with a textual representation of $G_t$
    (cf.\ \secref{sec:dataset}), we enhance each node $v_i$ with an additional
    field containing its visualized intermediate state. This creates a multimodal
    program where the model processes textual tokens interleaved with image
    patch embeddings (cf.\ \figref{fig:multimat-details}). To manage the
    increased context size from image embeddings, we omit node parameters
    (which are implicitly encoded in the visualizations) but explicitly include node
    output type information (e.g., grayscale or color) that the model cannot
    infer from the visualization alone.
  \item[Graph Conditioning] This approach more closely mirrors human visual
    experience by conditioning \projectname solely on a visualization
    of the entire graph $G_t$ with embedded intermediate visual outputs $I_t$, as shown in
    \figref{fig:multimat-details}. The model generates subsequent node
    $v_{t+1}$ using only this complete visual context, without explicit access
    to underlying textual node definitions.
\end{description}
At their core, both approaches remain autoregressive language models, and
\projectname can be trained by minimizing a cross-entropy objective:
\begin{equation}\label{eq:objective}
  \mathcal{L} = -\sum_{t=1}^{T}\sum_{s=1}^{S} \log p(\,v_{t,s} \mid v_{t,<s}, G_t, I_t, x; \theta\,),
\end{equation}
where $v_{t,s}$ is the expected token of $v_t$ in our intermediate text format
at position $s$, $v_{t,<s}$ represents previous tokens, $x$ denotes the input
conditions (which can be empty for unconditional generation), and $\theta$
represents the model parameters.

\subsection{Incremental Tree Search}\label{sec:tree-search}

Another advantage of topological node ordering is the ability to validate
node definitions incrementally during generation. By invoking our transpiler
and material engine at each step, we can detect syntactic and semantic errors
immediately rather than waiting until the entire graph is complete. When an
erroneous node definition is encountered, we execute an adaptive backtracking
strategy: first discarding and resampling the problematic node, and if errors
persist, inferring deeper structural issues by reversing further back in the
generation sequence. Specifically, we discard the $2^{(i-1)}$ most recently
generated nodes, where $i$ represents the current backtracking iteration.
This approach effectively transforms our generation process into an
\emph{incremental tree search} on a tree $\mathcal{T}$ of valid and invalid
nodes (cf.\ \figref{fig:tree-search}), systematically exploring the solution
space to discover valid programs. This incremental validation approach
identifies invalid outputs much faster than previous approaches, which require
sampling complete programs before validation can commence.
\begin{figure}
  \centering
  \begingroup %chktex-file 36

%https://tex.stackexchange.com/a/216086
\pgfdeclaredecoration{simple line}{initial}{
  \state{initial}[width=\pgfdecoratedpathlength-1sp]{\pgfmoveto{\pgfpointorigin}}
  \state{final}{\pgflineto{\pgfpointorigin}}
}

\tikzset{
  error/.style={fill=nord11,draw=nord11-dim},
  valid/.style={fill=nord14,draw=nord14-dim},
  active/.style={draw=nord0},
  line/.style={draw=nord6-dim,line width=.3mm},
  arrow/.style={line,->,>={Stealth[round,width=1.5mm,length=1mm]}},
  active arrow/.style={arrow,active,shorten <=.75ex,shorten >=.75ex,decorate,decoration={simple line,raise=#1}},
  snake it/.style={decorate,decoration={snake,post length=1mm}},
}

\newcommand{\valid}{\makebox(0,0){\textcolor{nord0}{\scriptsize\ding{51}}}}%
\newcommand{\invalid}{\makebox(0,0){\textcolor{nord0}{\scriptsize\ding{55}}}}%

\begin{forest}
  for tree={
    anchor=center,
    circle,
    line,
    fill=nord6,
    draw=nord6-dim,
    s sep*=2,
    l*=.5,
    edge path={%
      \noexpand\path[line,\forestoption{edge}] (!u.parent anchor)--(.child anchor);
      \noexpand\path[\forestoption{edge label}] (!u.parent anchor)--(.child anchor);
    }
  }
  [,phantom,s sep=1cm,
    % first
    [\valid,valid,active,name=first
      [\valid,valid,active,edge={active,densely dotted},edge label={active arrow=-1ex}
        [,phantom]
        [\invalid,error,active,edge={active,densely dotted},edge label={active arrow=1ex}]
      ]
      [\invalid,error]
      [\invalid,error]
    ]
    % second
    [\valid,valid,name=second
      [\valid,valid,active
        [\valid,valid,active,edge={active,densely dotted},edge label={active arrow=-1ex},name=valid-1]
        [\invalid,error,active,edge={active},edge label={active arrow=1ex,<-}]
      ]
      [\invalid,error]
      [\invalid,error]
    ]
    % third
    [\valid,valid,name=third
      [\valid,valid,active
        [\invalid,error,active,edge={active,densely dotted},edge label={active arrow=-1ex,}]
        [\invalid,error,active,edge={active},edge label={active arrow=1ex,<-}]
      ]
      [\invalid,error]
      [\invalid,error]
    ]
    % fourth
    [\valid,valid,active,name=fourth,calign=child,calign child=2
      [\valid,valid,active,edge={active},edge label={active arrow=-1ex,<-}
        [\invalid,error,active,edge={active},edge label={active arrow=-1ex,<-}]
        [\invalid,error]
      ]
      [\invalid,error]
      [\invalid,error]
      [\valid,valid,active,edge={active,densely dotted},edge label={active arrow=1ex}
        [\valid,valid,active,edge={active,densely dotted},edge label={active arrow=1ex},name=valid-2]
      ]
    ]
  ]
  \draw[arrow,active,densely dotted] (valid-1.south) to ($(valid-1.south)-(0,1.5ex)$);
  \draw[arrow,active,densely dotted] (valid-2.south) to ($(valid-2.south)-(0,1.5ex)$);
  \node[anchor=south] at ($(first)+(0,1ex)$) {\scriptsize(1)\strut};
  \node[anchor=south] at ($(second)+(0,1ex)$) {\scriptsize(2a\strut)};
  \node[anchor=south] at ($(third)+(0,1ex)$) {\scriptsize(2b\strut)};
  \node[anchor=south] at ($(fourth)+(0,1ex)$) {\scriptsize(3\strut)};
\end{forest}
\endgroup
  \caption{Visualization of our inference algorithm as a tree search. Tree
  nodes represent generated node definitions, and edges represent possible
  continuations. The algorithm proceeds as follows: generation continues until
  an invalid state (\xmark) is encountered (1), triggering backtracking to the
  previous node; from this point, if a valid node (\cmark) is generated, normal
  generation resumes (2a), but if invalid outputs persist (2b), the algorithm
  backtracks further until a valid path is found (3).}%
  \label{fig:tree-search}
\end{figure}

\subsection{Automatic Error Repair}\label{sec:auto-repair}
Through systematic analysis of failure cases, we identified recurring error
patterns that could be repaired automatically: (1) removal of extraneous
parameters that are specified for node types that do not support them, and (2)
automatic insertion of conversion nodes to resolve type mismatches between
connected nodes. For instance, when a color output is erroneously connected to
a grayscale input, we automatically insert an appropriate grayscale conversion
node. Conversely, when a grayscale output feeds into a color input, we insert a
gradient map node to perform type conversion. These repair mechanisms increase
the proportion of valid generations without requiring additional sampling
steps.

\section{Dataset}\label{sec:dataset}
\begin{wraptable}[1]{1}*
  \centering
  \begin{threeparttable}
    \begin{tabular}[c]{l s{1,3} d{3.0} cc}
      \toprule
      \thead{Models} & \nhead{Size} & \nhead{Max Nodes} & \nhead{Feature Set} & \nhead{Program}\\
      \midrule
      \matformer               & 2,820 & \dnote{3.0}{400} & Subset   & Designer\\
      \matformer*[Cond]         & 4,667 &  \dnote{3.0}{80} & Subset   & Designer\\
      \vlmaterial              & 3,663 &               30 & Limited  & Blender\\
      \undersmash{\projectname} & 6,878 &              128 & Complete & Designer\\
      \bottomrule
    \end{tabular}
    \begin{tablenotes}
      \item [1] Upper bound in complex filtering pipeline, actual could be less.
    \end{tablenotes}%
    \caption{Comparison of training data of
      \matformer~\citep{guerrero2022matformer}, conditional
      \matformer~\citep{hu2023materials},
      \vlmaterial~\citep{li2025vlmaterial}, and \projectname~(ours). We
      assembled the largest dataset with the most comprehensive set of
      features.}\label{tab:dataset-statistics}
  \end{threeparttable}
\end{wraptable}
To support the training and evaluation of \projectname, we collect procedural
materials from Adobe's Substance 3D Assets
Repository~\citep{SubstanceAssets2025}. Unlike previous work that either
focuses on basic graphs utilizing only a subset of Substance Designer
features~(e.g., lacking complex nodes such as pixel processors or function
graphs; \citealp{guerrero2022matformer,hu2023materials}) or targets other tools
with more limited capabilities~\citep{li2025vlmaterial}, our approach supports
the complete feature set. This comprehensive coverage enables us to collect
over 6\,000 unique materials, substantially more than existing datasets.
\tabref{tab:dataset-statistics} summarizes key characteristics of our dataset
compared to prior work.

\paragraph{Human-Readable Graph Representation}
Substance Designer's native file format (\sbs[]) has not been designed for
human readability, containing verbose XML structures, embedded binary data,
legacy metadata, and other implementation details, which makes direct language
modeling impractical. To address this, we develop a bidirectional transpiler
that converts between SBS and a compact, human-readable YAML-based
representation with topological node order, which we call \emph{\sbs}. Unlike
previous approaches that support only partial feature
sets~\citep{guerrero2022matformer,hu2023materials}, our transpiler preserves
the complete functionality of Substance graphs with programs that are, on
average, over 80\% shorter. Models operate exclusively in \sbs, with outputs
transpiled back to \sbs[] for execution. We provide representative examples in
\figref{fig:multimat-details} and complete program listings in
\appref{sec:additional-examples}.

\paragraph{Graph Preprocessing}
Our preprocessing pipeline standardizes graphs for the PBR workflow, focusing
on five essential texture maps: base color, normal, roughness, metallic, and
height. We trace backwards from these outputs to identify all contributing
nodes, pruning unconnected components and other output maps. Graphs containing
embedded bitmap graphics and SVGs are excluded to keep graphs fully procedural.
We further filter out graphs exceeding 128 nodes and flatten hierarchical
structures by inlining nested subgraphs and custom author dependencies into the
main graph. Non-atomic nodes from the standard Substance Designer library
remain as external references.

\section{Experiments}\label{sec:experiments}
We build \projectname models upon the \qwen vision-language model which
leverages a late fusion approach to combine image and text
tokens~\citep{bai2025qwen25vltechnicalreport}. We train and evaluate separate
models for unconditional generation (cf.\ \secref{sec:unconditional}) and
inverse procedural material synthesis (cf.\ \secref{sec:conditional}). We also
conduct a human evaluation (cf.\ \secref{sec:human}). Across all model
variants, we maintain a consistent maximum sequence length of 8\,192 tokens.
The training setup consists of 5 epochs using
\adam~\citep{loshchilov2018decoupled}, a learning rate of \lr{5}{5}, and a
batch size of 128. To ensure diversity in our generated outputs, we set the
inference sampling parameters to a temperature of 0.8 and a top-p value of
0.95. We provide examples in \figref{fig:examples} and
\appref{sec:additional-examples}. We ablate incremental tree search in
\secref{sec:analysis}.

\subsection{Evaluation of Unconditional Generation}\label{sec:unconditional}
For unconditional generation, the mixed conditioning variant,
\emph{\projectname[Mixed]}, embeds node previews at $140 \times 140$
resolution, resulting in 25 patch embeddings per image. For the graph
conditioning variant, \emph{\projectname[Graph]}, graph visualizations can utilize
up to 6\,144 tokens, with larger images downscaled to accommodate this limit.
We generate 100 outputs per model for evaluation.

\paragraph{Baselines}
For text-only procedural material synthesis, \vlmaterial represents the current
state-of-the-art approach. However, its Blender-specific training makes direct
comparison with our method difficult. We therefore create \vlmaterial[SBS] by
retraining a \vlmaterial-style model on our dataset for fair comparison. Unlike
the objective in equation~\eqref{eq:objective}, \vlmaterial is
trained to generate complete graphs in a single pass. However, during
inference, we can still validate nodes as they are generated and roll back upon
detecting irreparable errors (cf.\ \secref{sec:tree-search}) or repair them
after generation completes (cf.\ \secref{sec:auto-repair}). This means the
progression from \vlmaterial[SBS] to \projectname[Mixed] to \projectname[Graph]
represents a comparable, gradual shift from complete graph generation toward
iterative node generation. Since \vlmaterial[SBS] does not receive any images
in the unconditional setting, we base it on the larger and more powerful
text-only model \qwen[3 (8B\@; \citealp{yang2025qwen3})], giving it a slight
advantage over our models.
While graphics program synthesis research typically also benchmarks against
proprietary large language models such as \gpt~\citep{openai2024gpt4ocard} or
\claude~\citep{anthropic2025claude}, which have demonstrated competitive
performance in related
domains~\citep{belouadi2024automatikz,belouadi2024detikzify,belouadi2025tikzero,rodriguez2025starvector},
these models' unfamiliarity with \sbs and inability to produce valid \sbs[]
output preclude their inclusion as baselines.

\paragraph{Metrics} Our multimodal task permits diverse evaluation schemes
for automatic evaluation. To evaluate the \emph{visual quality} of generated
materials, we compute the Kernel Inception Distance (\thead{\kid};
\citealp{binkowski2018kid}), which compares the distribution of generated
material maps with material maps from our dataset. To detect
degenerate low \kid scores due to \emph{memorization} of training data (a
legitimate concern given our relatively small dataset), we also calculate
\thead{\rougel} scores~\citep{lin2004rouge} between the \sbs representation of
our generated materials and the training set (with masked parameters). This
metric computes the longest common subsequence and serves as an effective
memorization indicator \citep{hans2024be}. Notably, we specifically require
\emph{consecutive} subsequences due to \sbs's limited syntactic diversity,
which could otherwise produce misleading matches. To measure \emph{efficiency},
we introduce the Node Error Ratio (\thead{\ner}), defined as the average ratio
between discarded nodes and the total number of generated nodes.

\paragraph{Results}
\begin{wraptable}[1]{1}*
  \centering
  \begin{tabular}{l d{2.3} d{1.3} d{2.3}}
    \toprule
    \thead{Models} & \nhead{\kid\down} & \nhead{\rougel\down} & \nhead{\ner\down}\\
    \midrule
    \vlmaterial[SBS]    & 14.155 & 3.641 & \second{1.3}{14.846}\\
    \projectname[Mixed] & \second{2.3}{6.752} & \second{1.3}{2.195} & \first{2.3}{8.923}\\
    \undersmash{\projectname[Graph]} & \first{2.3}{2.365} & \first{1.3}{1.915} & 15.024\\
    \bottomrule
  \end{tabular}
  \caption{System-level $\text{scores}\times100$ for unconditional generation.
    Bold and underlined values indicate the best and second-best scores for
    each metric column, respectively. Arrows indicate metric directionality.
    \projectname[Graph] achieves the best overall performance.}%
  \label{tab:uncond-results}
\end{wraptable}
\begin{table}
  \centering

  \begin{tabular}{l *{2}{d{2.3}} d{1.3} d{2.3} d{1.3} d{2.3}}
    \toprule
    \thead{Models} & \nhead{\dsim\up} & \nhead{\clip\up} & \nhead{\style\down} & \nhead{\kid\down} & \nhead{\rougel\down} & \nhead{\ner\down}\\
    \midrule
    \vlmaterial[SBS]    & 31.344 & 65.678 & 3.211 & 14.976 & \first{1.3}{1.621} & \second{2.3}{16.933}\\
    \projectname[Mixed] & \second{2.3}{34.922} & \second{2.3}{66.737} & \second{1.3}{3.199} & \second{2.3}{3.675} & 2.194 & \first{2.3}{12.388}\\
    \undersmash{\projectname[Graph]} &  \first{2.3}{36.609} & \first{2.3}{67.907} & \first{1.3}{3.178} & \first{2.3}{2.801} & \second{1.3}{2.037} & 17.046\\
    \midrule
    \vlmaterial*[SBS]    & 31.348 & 65.867 & 3.126 & 27.862 & & \\
    \projectname*[Mixed] & \second{2.3}{40.258} & \second{2.3}{69.687} & \second{1.3}{3.093} & \second{2.3}{17.792}& & \\
    \undersmash{\projectname*[Graph]} & \first{2.3}{40.367} & \first{2.3}{70.114} & \first{1.3}{3.046}    & \first{2.3}{14.886}& & \\
    \bottomrule
  \end{tabular}
  \caption{System-level $\text{scores}\times100$ for conditional (inverse)
  generation, without (top) and with (bottom)  parameter optimization. Bold and
  underlined values indicate the best and second-best scores for each metric
  column, respectively. Arrows indicate metric directionality\@. \rougel and
  \ner scores remain unchanged by parameter optimization and are shown only
  once\@. \projectname[Graph] and \projectname*[Graph] achieve the best overall
  performance.}%
  \label{tab:cond-results}
\end{table}
\begin{figure}
  \resizeToWidth{\begin{tikzpicture}
    \pgfplotsset{
       diverging/.style={
        xbar stacked,
        enlarge y limits=.25,
        width=\pgfplotswidth,
        height=3.5cm,
        ytick={1,...,3},
        bar width=12.5pt,
        axis line style={draw=none},
        table/x expr={\thisrow{X}/8/33},
        xticklabel=\empty,%{\pgfmathparse{\tick*100}\pgfmathprintnumber{\pgfmathresult}\%},
        xmin=0,
        xmax=1,
      }
    }%

    \begin{axis}[
        diverging,
        yticklabels={\multimat*[Mixed],\vlmaterial*[SBS],\vlmaterial*[SBS]},
        ytick pos=right,
        axis y line*=right,
        axis x line=none,
      ]
      \addplot[draw=none] coordinates {(.5,1) (.5,2) (.5,3)};
    \end{axis}%
    \begin{axis}[
        diverging,
        nodes near coords={
          \pgfkeys{/pgf/fpu=true}
          \pgfmathparse{100*\pgfplotspointmeta}
          $\pgfmathprintnumber[fixed, precision=1]{\pgfmathresult}\%$
          \pgfkeys{/pgf/fpu=false}
        },
        nodes near coords custom/.style={
          every node near coord/.style={
            check for small/.code={
              \pgfkeys{/pgf/fpu=true}
              \pgfmathparse{\pgfplotspointmeta<#1}%
              \pgfkeys{/pgf/fpu=false}
            },
            check for small,
          },
        },
        nodes near coords custom=3,
        yticklabels={\multimat*[Graph],\multimat*[Graph],\multimat*[Mixed]},
        grid,
      ]
      % https://tex.stackexchange.com/a/442633
      \addplot table[row sep=\\] {
         X Y\\
       154 1\\ % graph-mixed
       214 2\\ % graph-sbs
       220 3\\ % mixed-sbs
      };
      \addplot table[row sep=\\] {
         X Y\\
       110 1\\
        50 2\\
        44 3\\
      };
    \end{axis}%
  \end{tikzpicture}}{\textwidth}\vspace{-\baselineskip}% FIXME: where does the empty line come from?
  \caption{Human preferences for model outputs as a diverging bar
  chart\@. \projectname*[Graph] is the most preferred model overall, while
  \vlmaterial*[SBS] is consistently the least
  preferred.}\label{fig:human-results}
\end{figure}
\tabref{tab:uncond-results} presents the system-level metric scores for our
evaluation\@. \projectname[Graph] leads in visual quality with the lowest \kid
score, outperforming \projectname[Mixed] by over \pp{4} (\pp) and
\vlmaterial[SBS] by more than \pp{11}. This considerable gap in performance
suggests that the better the visual representations are aligned with human
creative workflows, the better the results---an intuitive but important
finding. All models exhibit minimal memorization, with \rougel scores showing
that no more than 4\% of any generated sequence matches a contiguous segment
from the training data. Nonetheless, both \projectname variants demonstrate
approximately \pp{1.5} lower copying rates compared to \vlmaterial[SBS],
suggesting slightly better generalization. Regarding efficiency,
\projectname[Mixed] excels with the lowest \ner, achieving a \pp{6} improvement
over the other models. Both \projectname[Graph] and \vlmaterial[SBS] show
comparable \ner scores around 15\%. For \projectname[Graph], these errors are
primarily due to OCR-like errors in reading node names and function types
embedded as text in graph images. In contrast, we attribute the errors in
\vlmaterial[SBS] to fundamental difficulties in understanding graph structures.
Despite these limitations, the error rates remain within acceptable bounds for
practical applications, and \projectname[Graph] emerges as the best overall
model.

\subsection{Evaluation of Conditional Generation}\label{sec:conditional}
As in prior work~\citep{hu2023materials,li2025vlmaterial}, we train inverse
\projectname variants that learn to generate procedural materials from rendered
images. These models follow the same training procedure as their unconditional
counterparts, with one key modification: each training example is preceded by a
$512 \times 512$ rendering of itself, which adds 324 additional image patches
to the model context. During inference, the model takes an image as input and
generates a corresponding procedural material. We reserve 100 examples from our
data as held-out test data for evaluation.

\paragraph{Baselines}
Analogously to \secref{sec:unconditional}, we adapt \vlmaterial for inverse
rendering with \sbs[] and use it as a baseline. Since an image input is now
required for \vlmaterial[SBS], we also base it on \qwen instead of \qwen[3
(8B)] and train it using the same method as \projectname.

\paragraph{Parameter Optimization}
To further refine generated materials, we apply gradient-based optimization
using differentiable rendering. This approach has proven effective for optimal
parameter
estimation~\citep{shi2020match,hu2022optim,li2023end,hu2023materials}. We
employ \diffmat~\citep{shi2020match,li2023end}, a widely adopted differentiable
renderer for Designer materials, to optimize the generated graphs against the
input images. Models using this refinement step are denoted as \projectname*
and \vlmaterial*[][,] respectively.

\paragraph{Metrics}
In addition to the metrics from \secref{sec:unconditional}, we evaluate
reconstruction quality by rendering the generated materials and comparing them
to the input images using perceptual similarity metrics. Specifically,
we measure cosine similarity between \thead{\clip} image
embeddings~\citep{radford2021clip,hessel2021clipscore}, compute \style* loss
(\thead{\style}; \citealp{gatys2016styleloss}) as the L1 distance between Gram
matrices of VGG features, and calculate \dsim*~(\thead{\dsim};
\citealp{fu2023dreamsim}), a learned perceptual similarity metric designed to
align with human judgments.

\paragraph{Results} \tabref{tab:cond-results} presents the system-level metric
scores for conditional evaluation. The perceptual similarity metrics
consistently demonstrate that \projectname[Graph] achieves the highest fidelity
to input images, with \projectname[Mixed] performing second-best and
\vlmaterial[SBS] ranking last. For example, \dsim* scores are 36.609, 34.922,
and 31.344, respectively, a ranking that mirrors our unconditional evaluation
results. Parameter optimization yields substantial improvements in perceptual
similarity, with \projectname*[Graph] and \projectname*[Mixed] showing average
gains of 6\% and 8\%, respectively. In contrast, \vlmaterial*[SBS] exhibits
minimal improvement (only 1\%), suggesting its outputs deviate too
far from the input for parameter optimization to be effective. Interestingly, while
parameter optimization improves perceptual similarity, \kid scores increase. This
could occur because optimization aligns outputs more closely with the test set,
which represents only a subset of the training distribution, potentially
increasing distance from the full distribution. Nevertheless, both \projectname
and \projectname* variants outperform \vlmaterial[SBS] and \vlmaterial*[SBS] on
\kid by over \pp{10}, respectively. The remaining metrics reinforce trends from
unconditional evaluation\@. \rougel scores do not  exceed 2\% (indicating
minimal memorization), and \projectname[Mixed] produces the fewest errors.
Overall, \projectname[Graph] and its optimized variant, \projectname*[Graph],
deliver the strongest performance across metrics.

\subsection{Human Evaluation}\label{sec:human}
To corroborate our automatic evaluation results, we conduct a human evaluation.
We employ comparative annotation~\citep{thurstone1927comp} and focus on the image
reconstruction/inverse rendering use case, which allows for intuitive human
assessment (cf.\ \figref{fig:examples}). Annotators receive triplets of
rendered generated materials from \vlmaterial*[SBS], \projectname[Mixed], and
\projectname[Graph] and identify which output best and least resembles the
input image. Following \citet{hu2023materials,li2025vlmaterial}, we generate
multiple programs ($N=40$) per model and image, selecting the result with
the highest \dsim* score as the final candidate. We test 33 input materials
from graphs filtered during preprocessing (e.g., due to excessive length),
which represent particularly challenging cases. Eight expert annotators with
extensive procedural material experience assess each triplet in randomized
order. As shown in \figref{fig:human-results}, annotators rank
\vlmaterial*[SBS] considerably lower than \projectname*[Mixed] and
\projectname*[Graph], and prefer \projectname*[Graph] over
\projectname*[Mixed]. These findings align with our automatic evaluation
rankings and demonstrate our approach's effectiveness in generating
perceptually similar materials.

\section{Analysis \& Discussion}\label{sec:analysis}
\begin{figure}[!t]
  \newlength\q
  \pgfmathsetlength{\q}{\textwidth/4 -2\tabcolsep}
  % https://tex.stackexchange.com/a/7318
  \begin{tabular}{>{\centering\arraybackslash}m{\q}|*{3}{>{\centering\arraybackslash}m{\q}}}
    \toprule
    \thead{Input} & \nhead{\clap{\vlmaterial*[SBS]\unskip}} & \thead{\projectname*[Mixed]} & \nhead{\clap{\undersmash{\projectname*[Graph]}}}\\
    \midrule
    \includegraphics[width=\q]                    {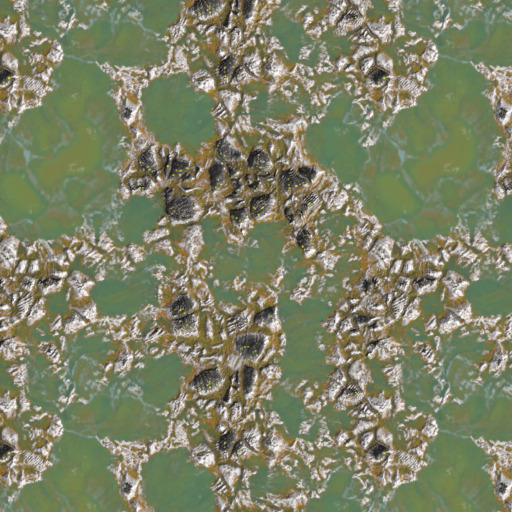} &
    \includegraphics[width=\q]{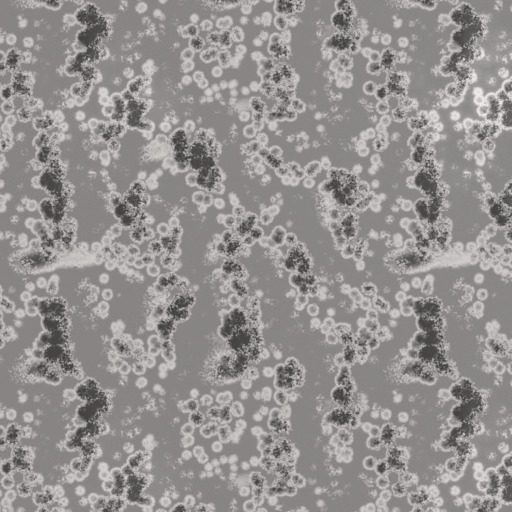} &
    \includegraphics[width=\q]{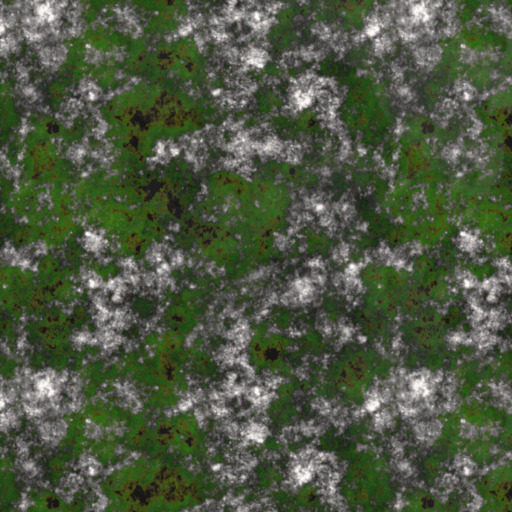} &
    \includegraphics[width=\q]{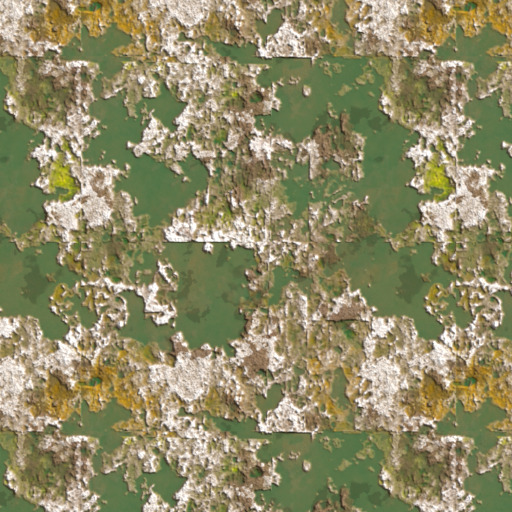}\\
    % \midrule
    \includegraphics[width=\q]                    {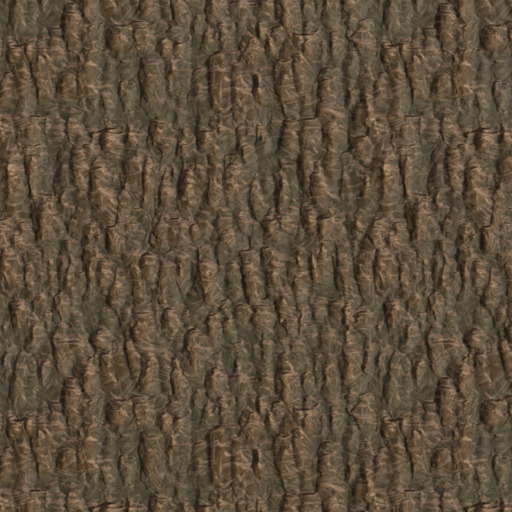} &
    \includegraphics[width=\q]{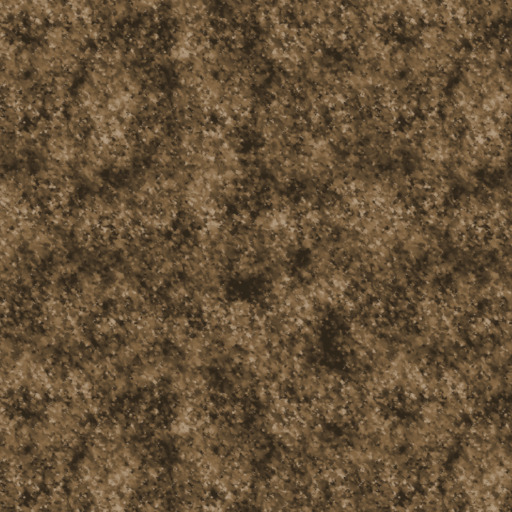} &
    \includegraphics[width=\q]{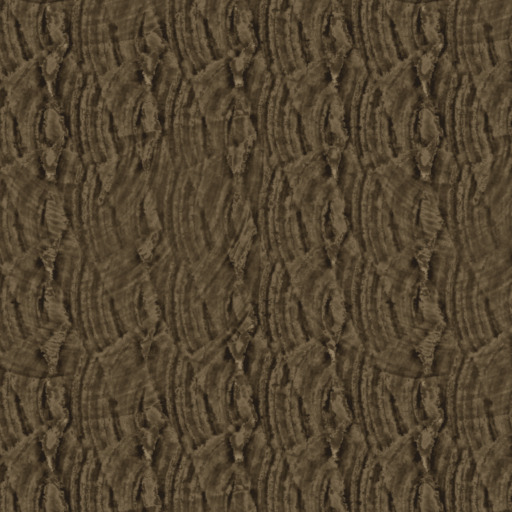} &
    \includegraphics[width=\q]{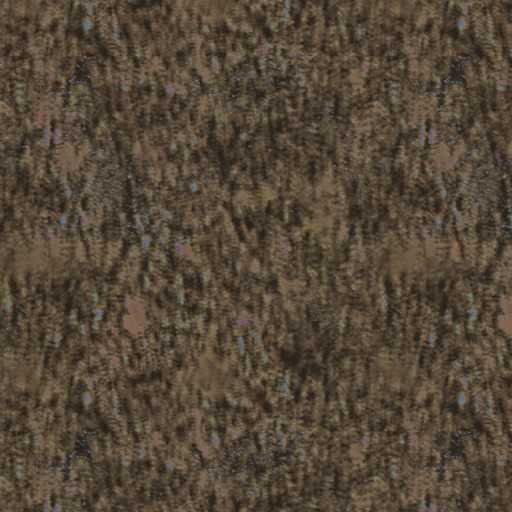}\\
    % \midrule
    \includegraphics[width=\q]                    {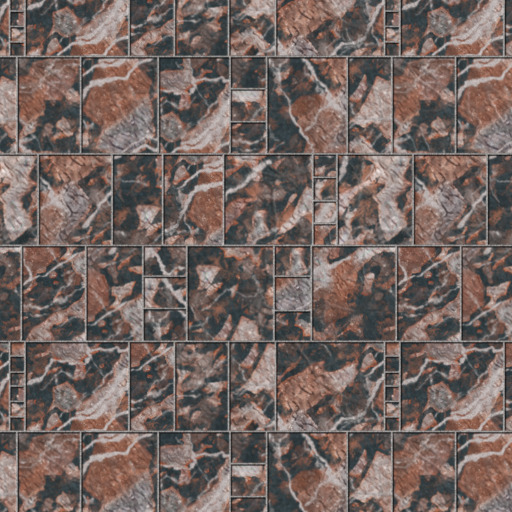} &
    \includegraphics[width=\q]{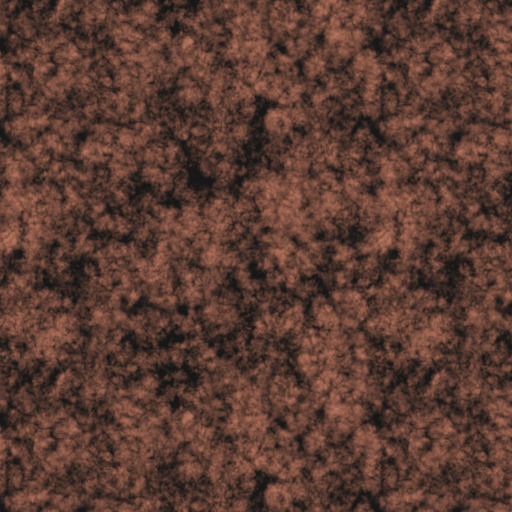} &
    \includegraphics[width=\q]{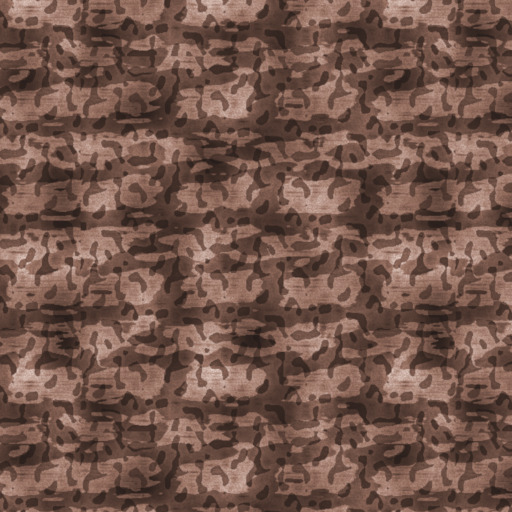} &
    \includegraphics[width=\q]{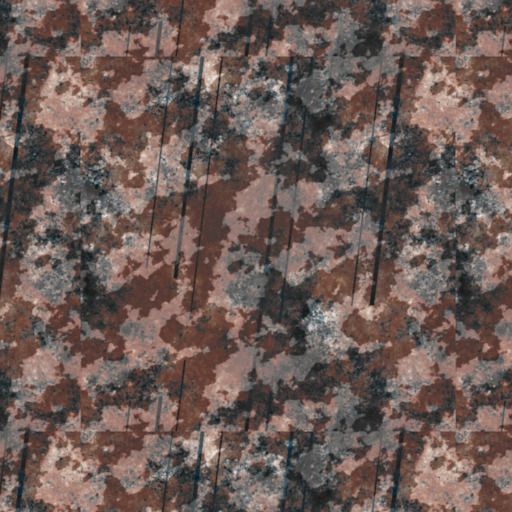}\\
    % \midrule
    \includegraphics[width=\q]                    {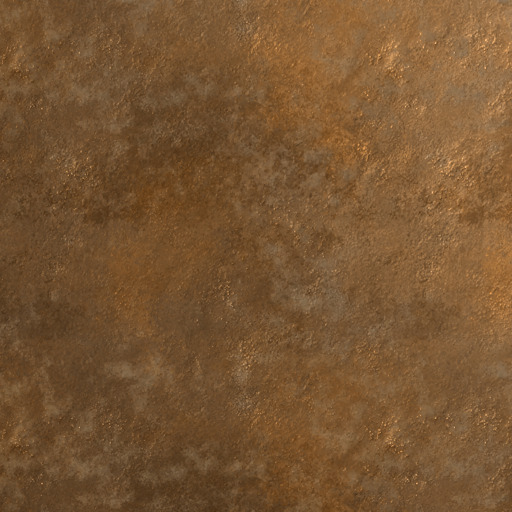} &
    \includegraphics[width=\q]{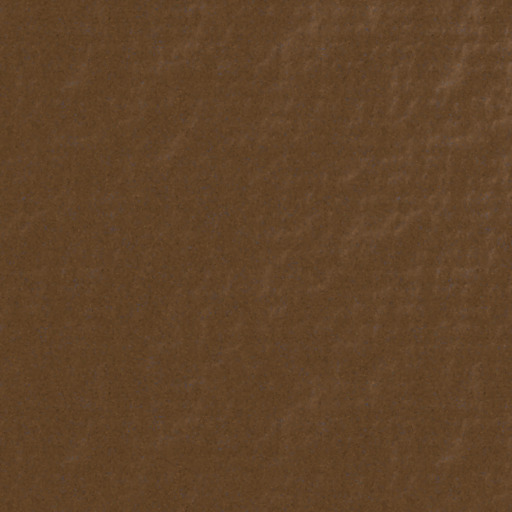} &
    \includegraphics[width=\q]{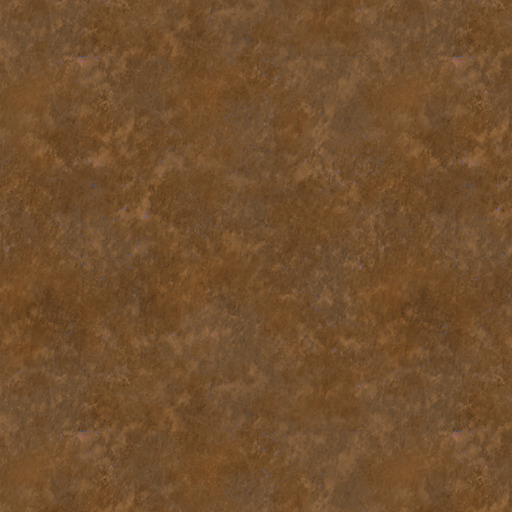} &
    \includegraphics[width=\q]{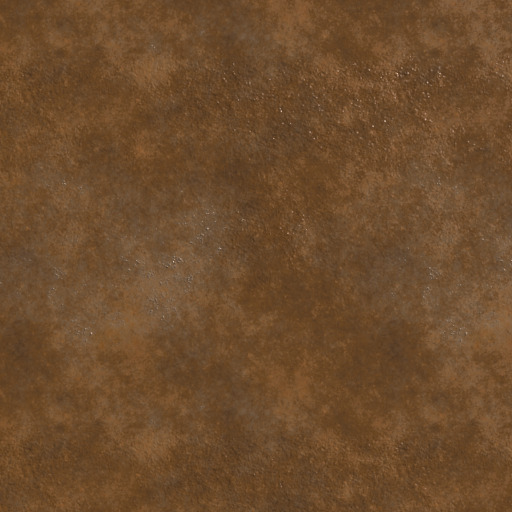}\\
    \bottomrule
  \end{tabular}
  \caption{Qualitative examples for inverse procedural material modeling
  following the setup of our human evaluation in \secref{sec:human}. The
  leftmost column shows input materials from graphs filtered during
  preprocessing, making these particularly
  challenging test cases\@. \projectname*[Mixed] consistently outperforms
  \vlmaterial*[SBS], while \projectname*[Graph] achieves the best results
  overall. Additional examples, including failure cases, are provided in
  \appref{sec:additional-examples}.}%
  \label{fig:examples}
\end{figure}
Our comparisons demonstrate that model performance improves steadily
as the degree of graph visualization increases, with \multimat[Graph] achieving
the highest performance overall~(cf.\
\tabtabref{tab:uncond-results}{tab:cond-results}; \figref{fig:human-results}).
 This finding aligns with how humans interact with
procedural materials---through visual node graph interfaces---and validates
established UX design principles in this domain.
\begin{wraptable}[1]{r}*
  \centering
  \begin{tabular}[c]{l d{1.2} d{2.2}}
    \toprule
    \thead{Models} & \nhead{Deletion\down} & \nhead{Conversion\down}\\
    \midrule
    \vlmaterial[SBS]    & 2.71 & 12.26\\
    \projectname[Mixed] & \second{1.2}{1.18} &  \first{2.2}{3.51}\\
    \projectname[Graph] & \first{1.2}{1.1} & \second{2.2}{6.49}\\
    \bottomrule
  \end{tabular}
  \caption{Percentage of nodes repaired through parameter deletion or
    conversion node insertion in our unconditional and conditional evaluations.
    Bold and underlined values indicate the best and second-best scores for
    each metric column, respectively. Arrows indicate metric directionality.
    Our \projectname models require the least amount of repair.}%
  \label{tab:error-statistics}
\end{wraptable}
The qualitative examples in \figref{fig:examples} further illustrate this
trend, with \vlmaterial*[SBS] struggling to generate faithful outputs,
indicating that purely text-based approaches are not ideal for expressive node
graph systems like Designer. This limitation persists even with more powerful
base models, as our unconditional generation experiments confirm.
Beyond architectural improvements, our tree search algorithm enables more
efficient graph generation; without it, models may have to resort to sampling
complete outputs for validation (the inference approach used by previous
methods), which is expensive. For instance, disabling tree search causes \ner
of \vlmaterial[SBS] to deteriorate further from 14.846 to 33.953, highlighting
how our search strategy can improve inference without further training.
The impact of automatic error repair is more nuanced, as shown in
\tabref{tab:error-statistics}. Only approximately 1\% of nodes generated by
\multimat contain hallucinated parameters, and fewer than 6.5\% require
conversion. In contrast, \vlmaterial exhibits nearly double the scores for both
repair mechanisms. This difference demonstrates that \vlmaterial requires
considerably more repair than our models and supports our claim that our models
possess a better understanding of graph structures. Notably, since corrections
are not fed back to the models, these results reflect their intrinsic
generation capabilities.

\section{Conclusion}
We present \projectname, a multimodal program synthesis framework and model
suite that generates procedural materials by incorporating visual feedback
throughout the generation process. Our key insight is that procedural material
graphs are inherently visual-spatial programs, and treating them as such leads
to substantial improvements over text-only approaches. By conditioning on
visual intermediate states---either interleaved with text (mixed conditioning)
or as complete graph visualizations (graph conditioning)---our models achieve
consistent improvements over text-only baselines. Our incremental tree search
algorithm further enhances generation efficiency by validating nodes as they
are created and backtracking upon errors. While we demonstrate \projectname
specifically for procedural material synthesis, we hope its general principles
will inspire further research at the intersection of computer graphics, program
synthesis, and multimodal AI\@.

\paragraph{Future Work}
The development of procedural material graph synthesis approaches is currently
constrained by limited training data availability. We plan to address this
challenge through self-learning
techniques~\citep{He2020RevisitingSF,Wei2020TheoreticalAO} that leverage our
unconditional models to generate synthetic supervised training data by
rendering outputs and subsequently training conditional models on this expanded
data. Additionally, we aim to develop a unified model trained across multiple
node graph systems to investigate potential transfer learning
benefits~\citep{pan2010transfer}. Beyond methodological advances, our
models offer promising practical applications: conditional models could extract
material graphs directly from photographic regions, while unconditional models
could power intelligent auto-completion features in user interfaces.
Furthermore, our methodology naturally extends to related domains such as
vector graphics
synthesis~\citep{wu2023iconshop,polaczek2025neuralsvg,rodriguez2025starvector,yang2025omnisvg},
where visual editing interfaces are similarly prevalent.

\paragraph{Limitations}
Although our models and baselines use the same or similar base models, they
generate graphs in fundamentally different ways, resulting in considerable
differences in training efficiency. Text-only models like \vlmaterial can
process entire graphs as single training examples, whereas \projectname must
adapt the visual context for each individual node, effectively processing
training examples one node at a time. This difference leads to much longer
training times: while \vlmaterial completes training in a few hours on
$8\times\text{A100}$ 80GB GPUs, \projectname models require several days on the
same hardware despite being trained on a comparable number of tokens.
Nevertheless, since the amount of procedural materials is very small
(regardless of the dataset), training times remain within acceptable bounds in
absolute terms, despite the relative differences between methods. Additionally,
both approaches achieve a more similar throughput during inference.

\section*{Ethics Statement}
We ensure that all procedural materials collected for model training are 
properly licensed and explicitly permit such usage, thereby preventing any 
copyright infringement. In adherence to this principle, we specifically 
exclude Substance 3D Community Assets~\citep{SubstanceCommunityAssets2025} 
from our training data due to licensing restrictions. While we acknowledge 
the use of generative models in preparing this manuscript, their application 
is strictly limited to writing assistance, such as paraphrasing, spell checking, 
and synonym suggestions.

\anonymize[]{\section*{Acknowledgements}%
  We thank the Adobe Substance 3D team for providing access to the Substance 3D
  Assets Repository and the Substance Automation Python API\@. We also thank
  our annotators for their valuable time. This work was conducted while the
  first author was an intern at Adobe Research, France.%
}

\bibliography{multimat}
\bibliographystyle{styles/iclr2026_conference}

\clearpage\appendix\FloatBarrier
\section{Additional Examples}\label{sec:additional-examples}
\begin{figure}
  \pgfmathsetlength{\q}{\textwidth/4 -2\tabcolsep}
  \begin{tabular}{>{\centering\arraybackslash}m{\q}|*{3}{>{\centering\arraybackslash}m{\q}}}
    \toprule
    \thead{Input} & \nhead{\clap{\vlmaterial*[SBS]\unskip}} & \thead{\projectname*[Mixed]} & \nhead{\clap{\undersmash{\projectname*[Graph]}}}\\
    \midrule
    \includegraphics[width=\q]                    {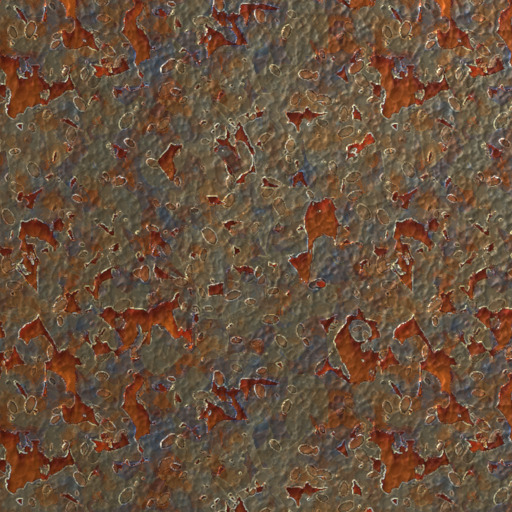} &
    \includegraphics[width=\q]{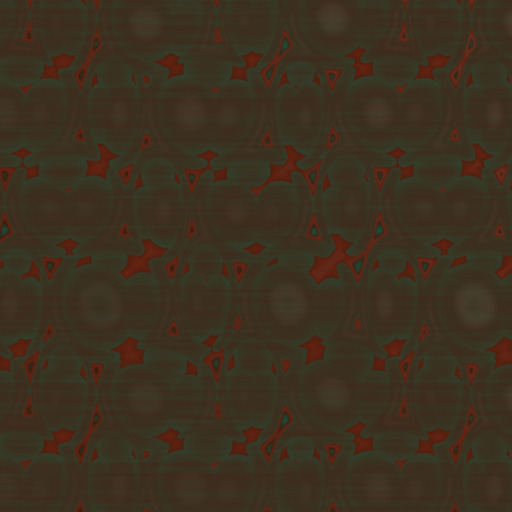} &
    \includegraphics[width=\q]{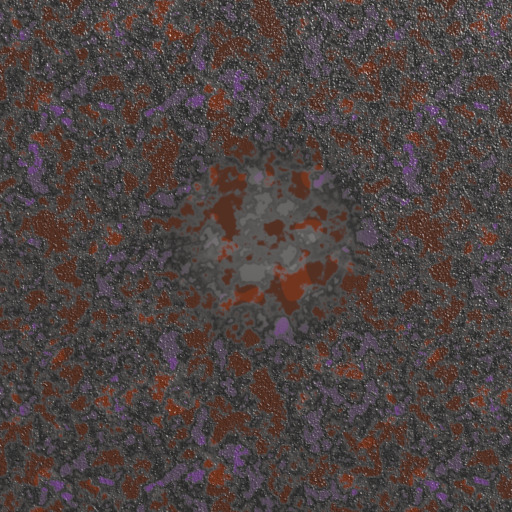} &
    \includegraphics[width=\q]{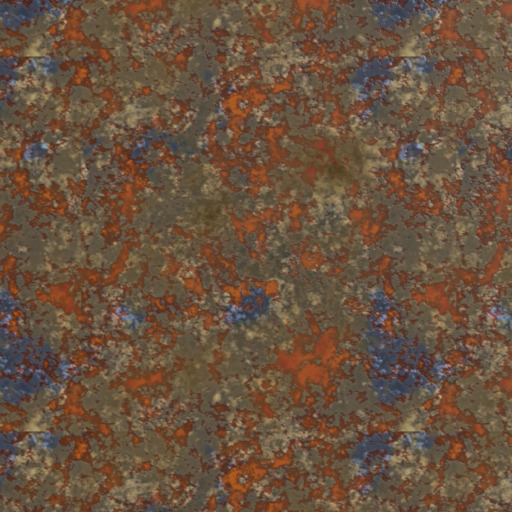}\\
    % \midrule
    \includegraphics[width=\q]                    {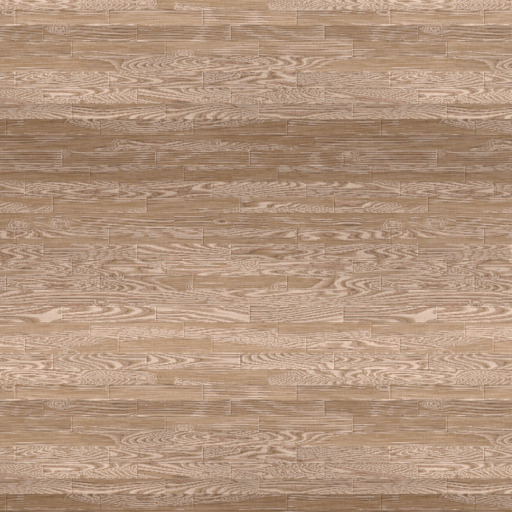} &
    \includegraphics[width=\q]{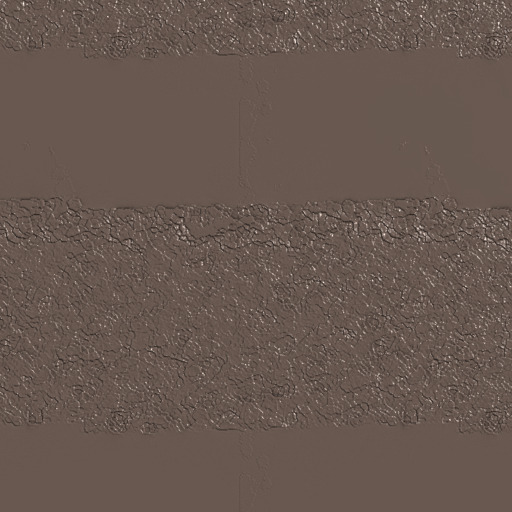} &
    \includegraphics[width=\q]{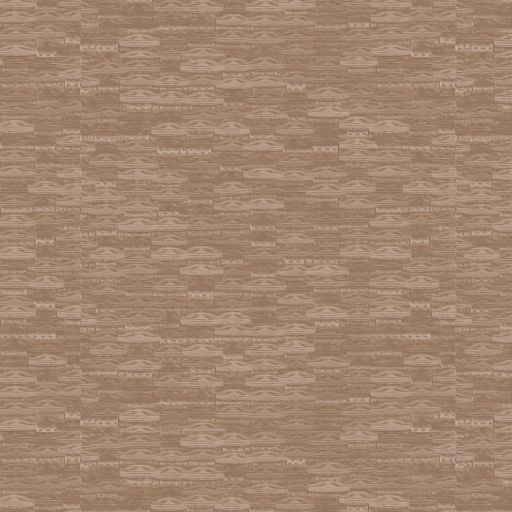} &
    \includegraphics[width=\q]{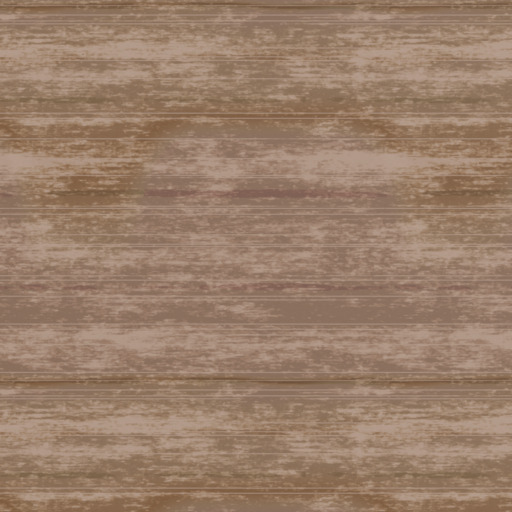}\\
    % \midrule
    \includegraphics[width=\q]                    {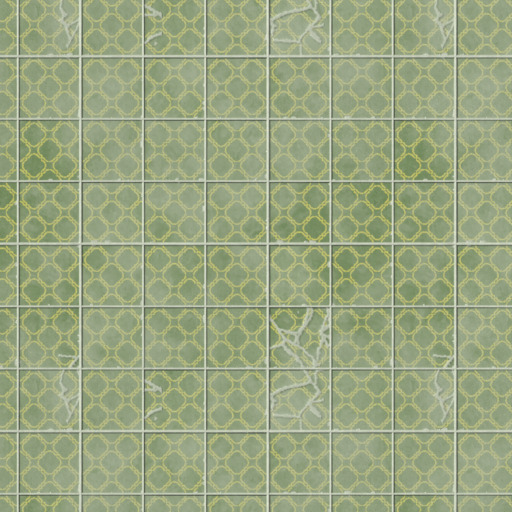} &
    \includegraphics[width=\q]{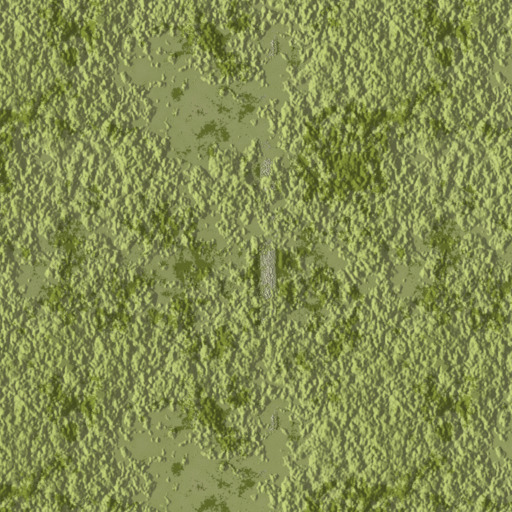} &
    \includegraphics[width=\q]{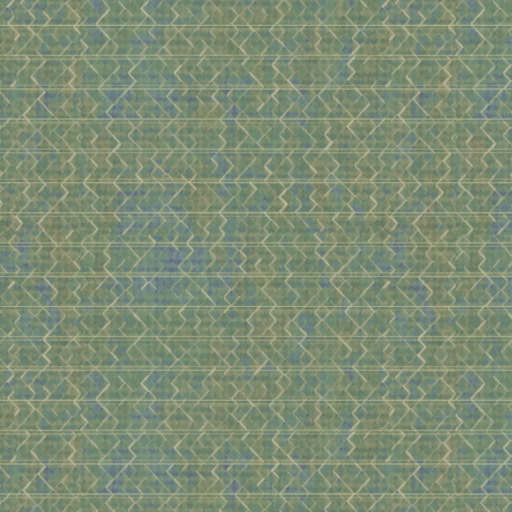} &
    \includegraphics[width=\q]{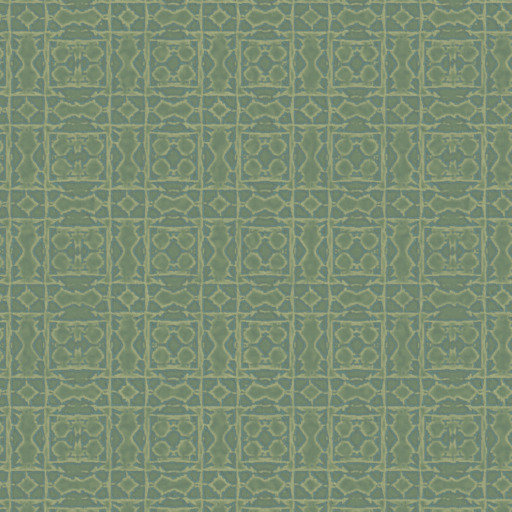}\\
    % \midrule
    \includegraphics[width=\q]                    {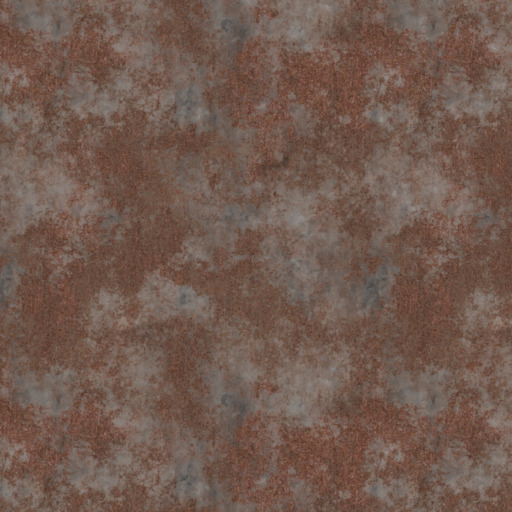} &
    \includegraphics[width=\q]{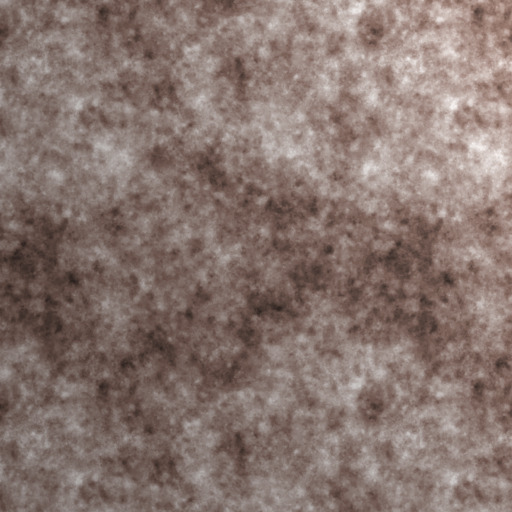} &
    \includegraphics[width=\q]{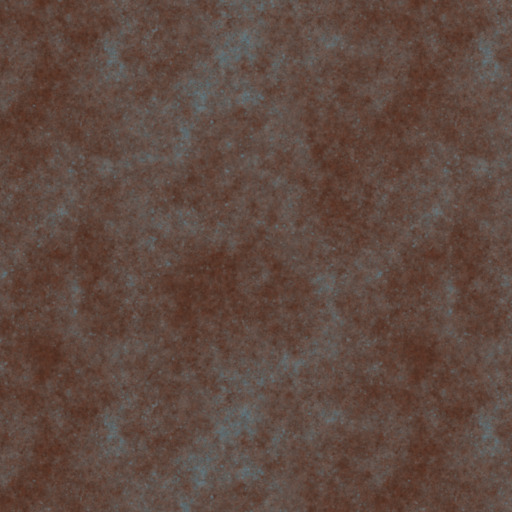} &
    \includegraphics[width=\q]{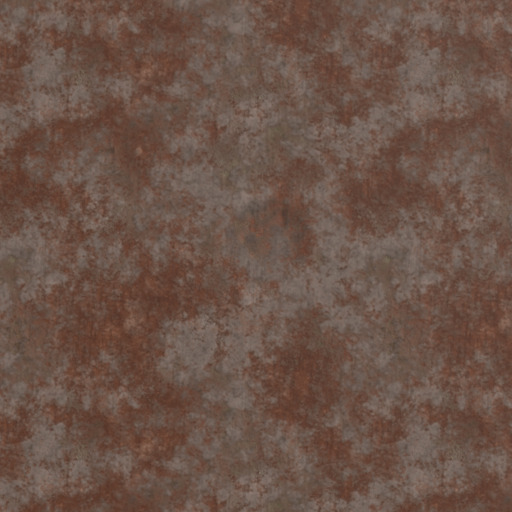}\\
    % \midrule
    \includegraphics[width=\q]                    {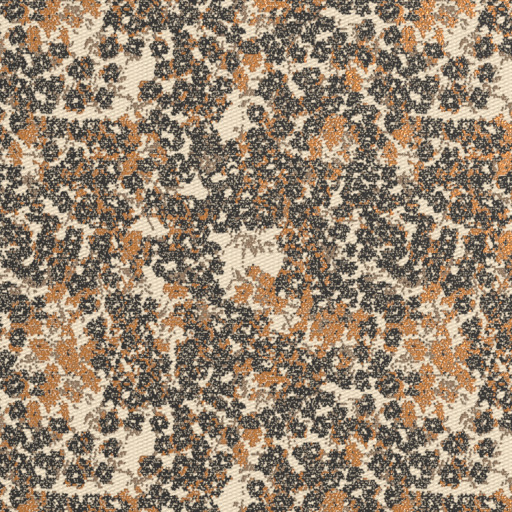} &
    \includegraphics[width=\q]{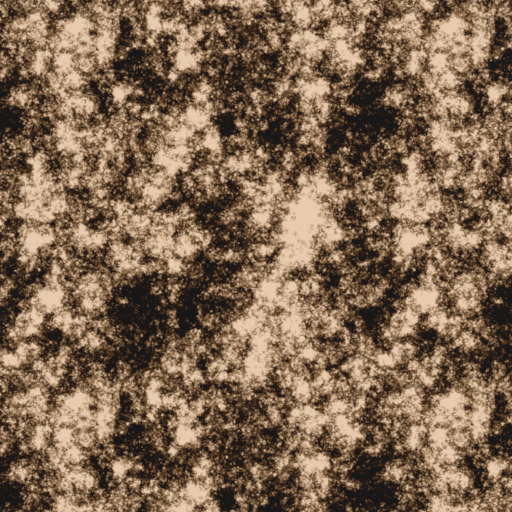} &
    \includegraphics[width=\q]{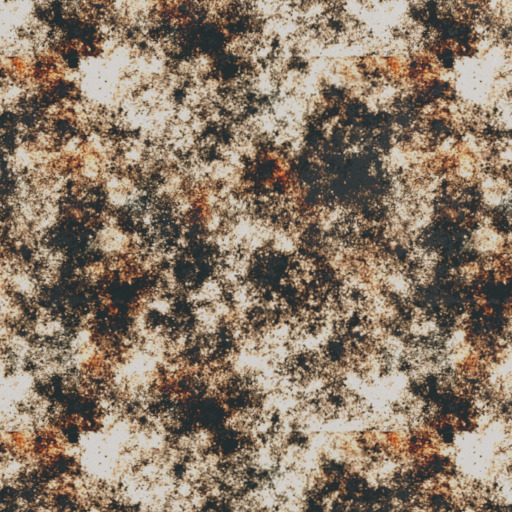} &
    \includegraphics[width=\q]{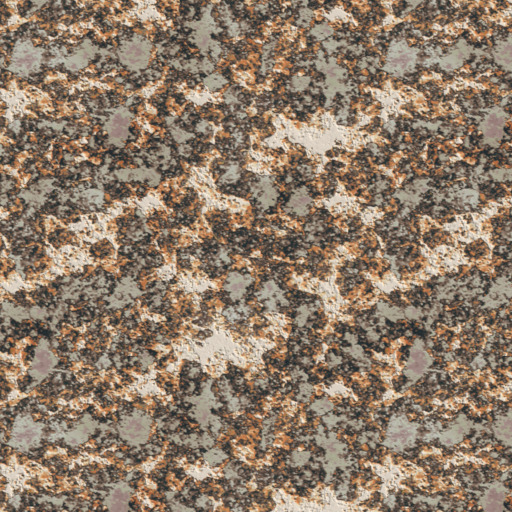}\\
    % \midrule
    \includegraphics[width=\q]                    {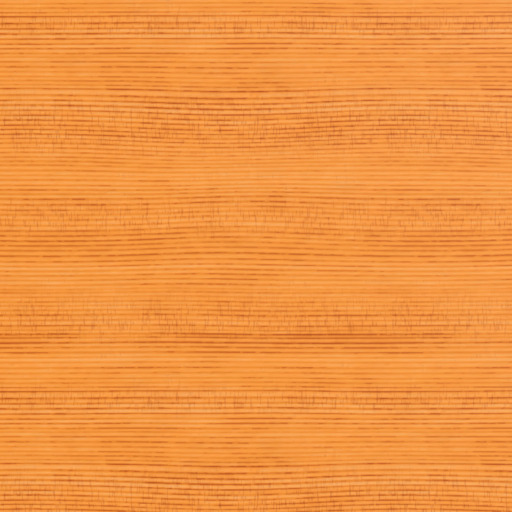} &
    \includegraphics[width=\q]{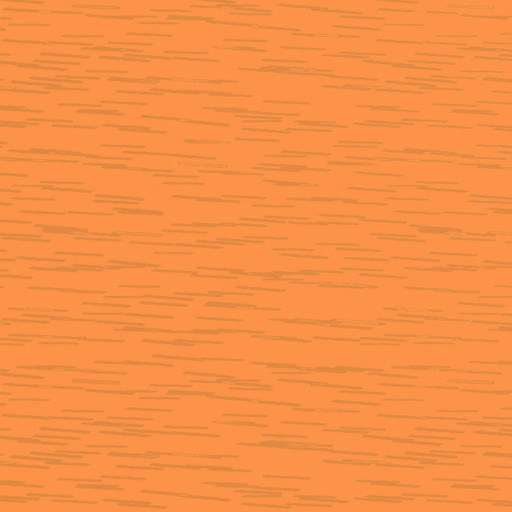} &
    \includegraphics[width=\q]{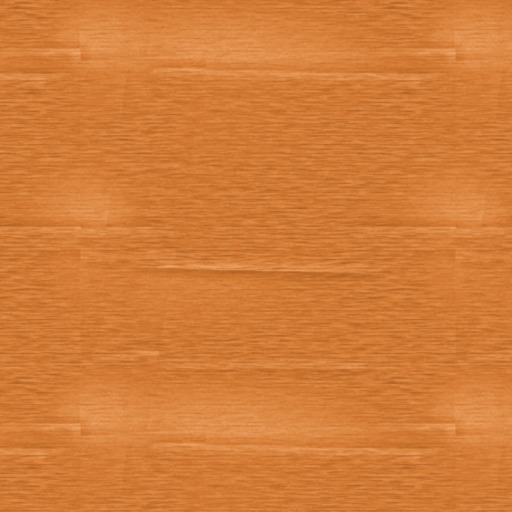} &
    \includegraphics[width=\q]{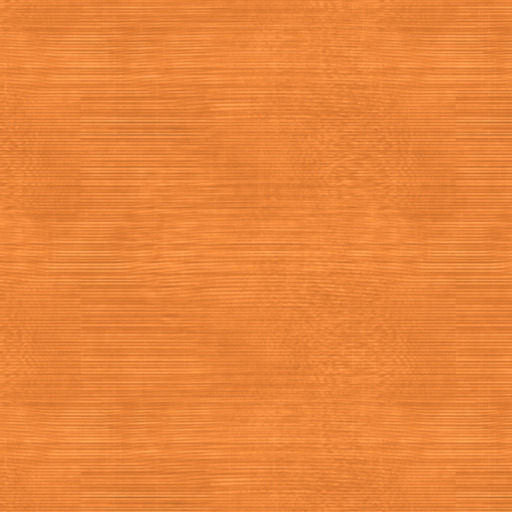}\\
    \bottomrule
  \end{tabular}
  \caption{Additional qualitative examples for inverse procedural material
  modeling following the setup of our human evaluation in \secref{sec:human}.
  The leftmost column shows input materials from graphs filtered during
  preprocessing, making these particularly challenging test cases\@.
  \projectname*[Mixed] consistently outperforms \vlmaterial*[SBS], while
  \projectname*[Graph] achieves the best results overall.}%
  \label{fig:more-examples}
\end{figure}

\begin{figure}
  \pgfmathsetlength{\q}{\textwidth/4 -2\tabcolsep}
  \begin{tabular}{>{\centering\arraybackslash}m{\q}|*{3}{>{\centering\arraybackslash}m{\q}}}
    \toprule
    \thead{Input} & \nhead{\clap{\vlmaterial*[SBS]\unskip}} & \thead{\projectname*[Mixed]} & \nhead{\projectname*[Graph]}\\
    \midrule
    \includegraphics[width=\q]                    {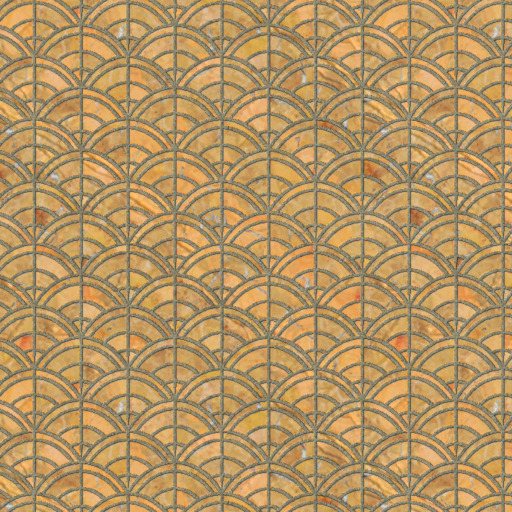} &
    \includegraphics[width=\q]{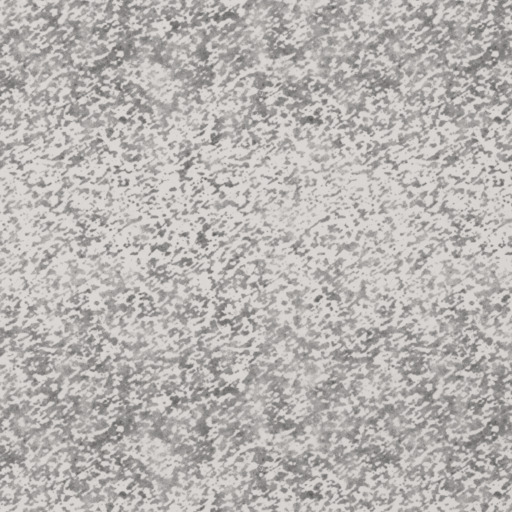} &
    \includegraphics[width=\q]{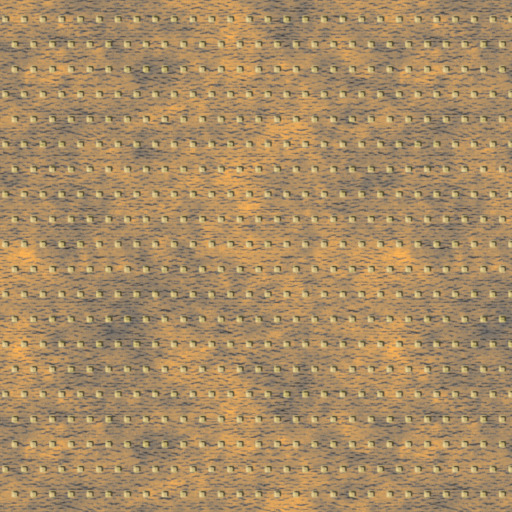} &
    \includegraphics[width=\q]{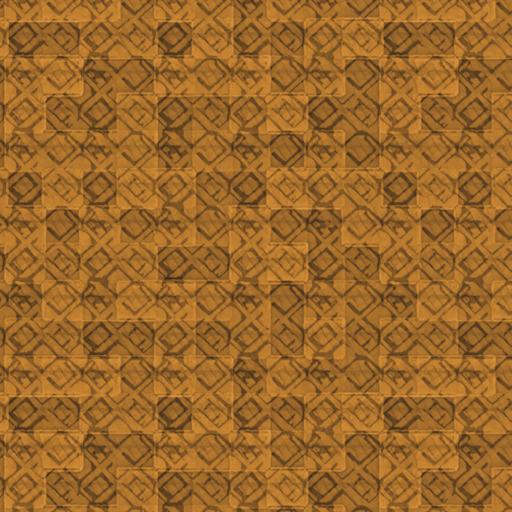}\\
    % \midrule
    \includegraphics[width=\q]                    {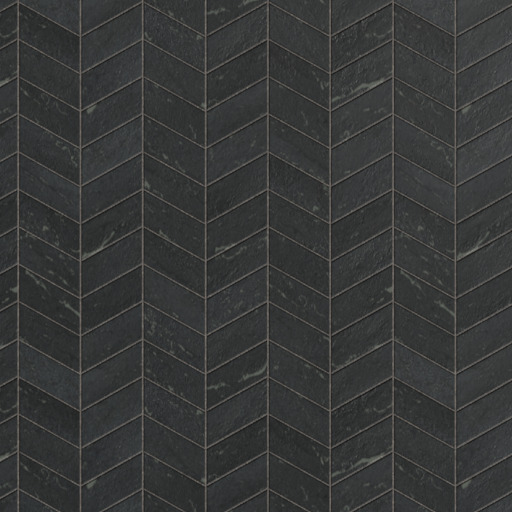} &
    \includegraphics[width=\q]{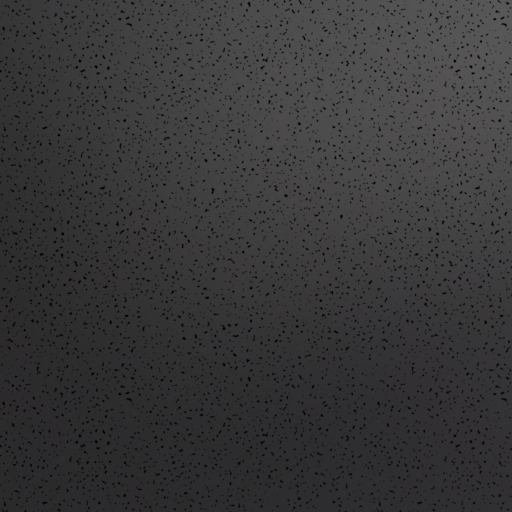} &
    \includegraphics[width=\q]{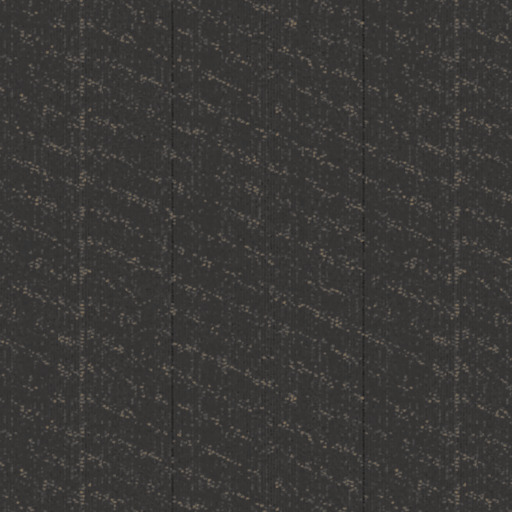} &
    \includegraphics[width=\q]{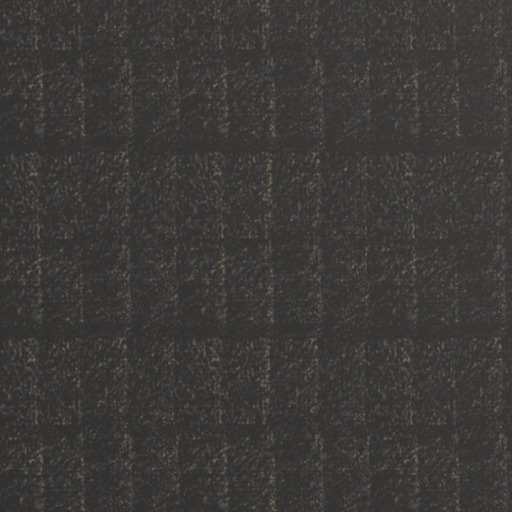}\\
    % \midrule
    \includegraphics[width=\q]                    {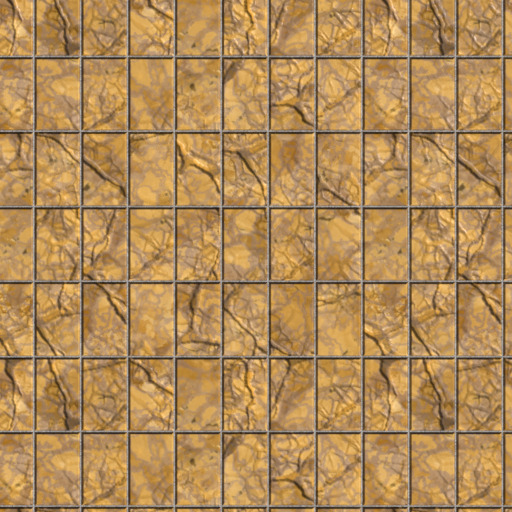} &
    \includegraphics[width=\q]{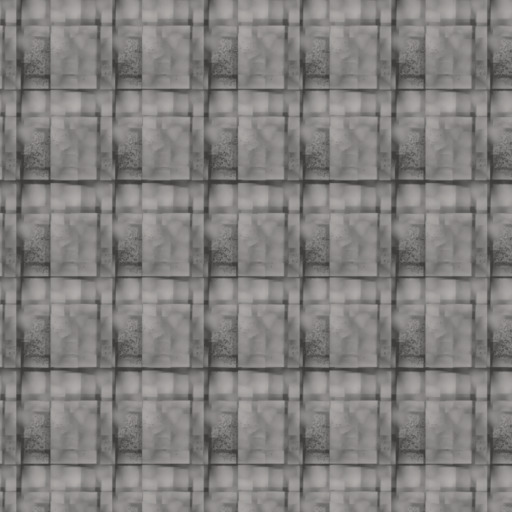} &
    \includegraphics[width=\q]{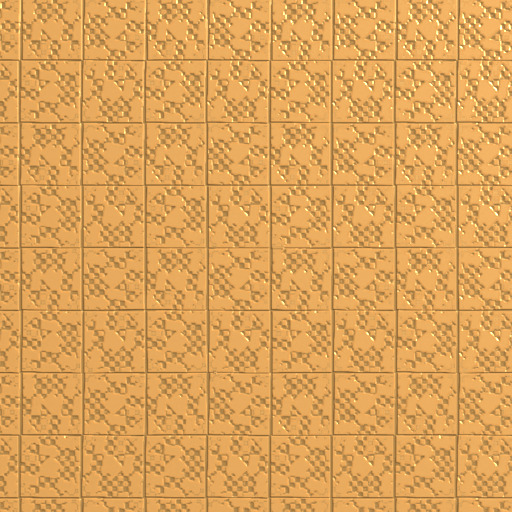} &
    \includegraphics[width=\q]{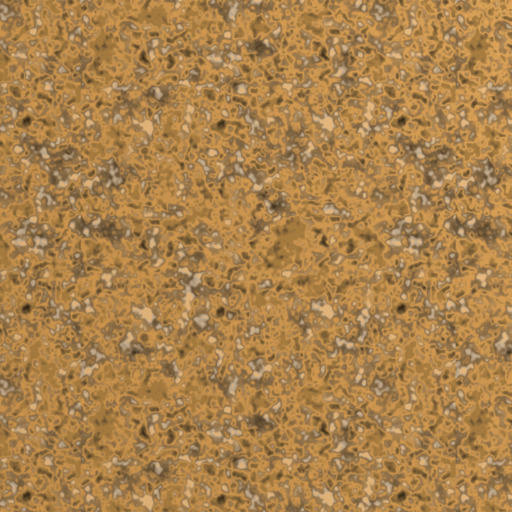}\\
    % \midrule
    \includegraphics[width=\q]                    {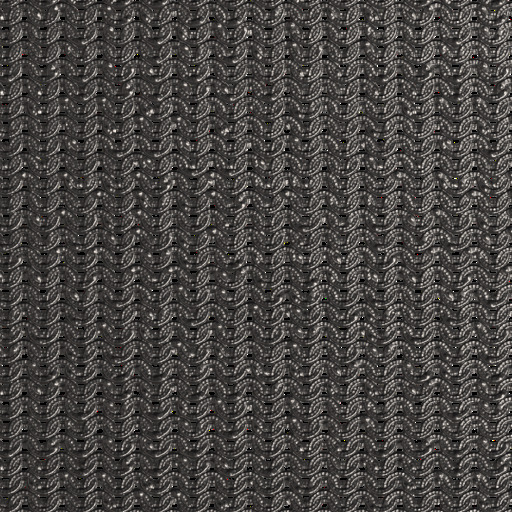} &
    \includegraphics[width=\q]{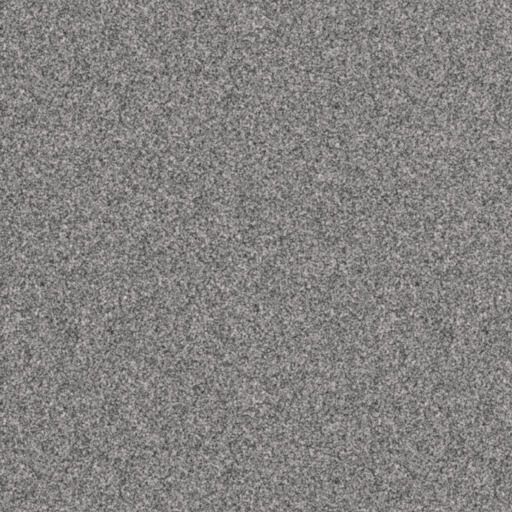} &
    \includegraphics[width=\q]{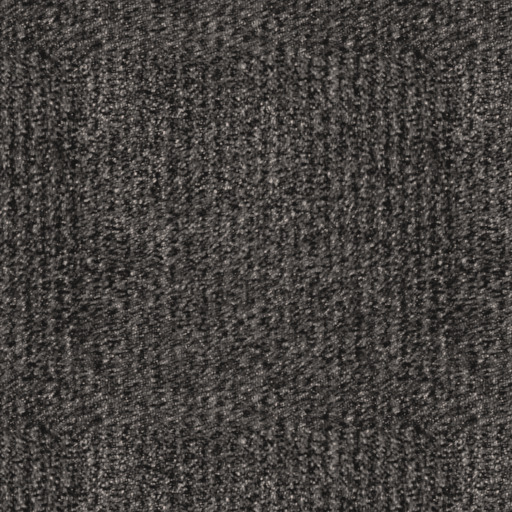} &
    \includegraphics[width=\q]{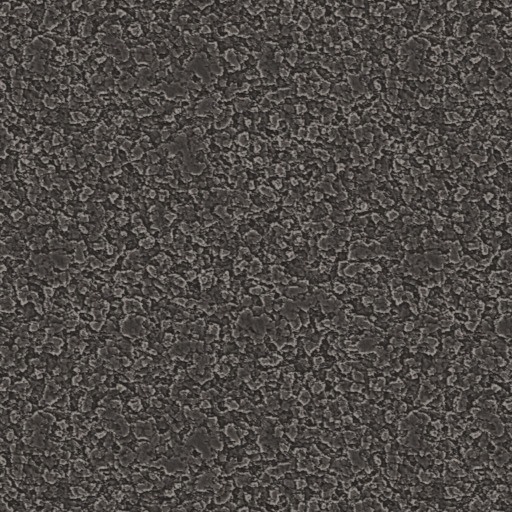}\\
    % \midrule
    \includegraphics[width=\q]                    {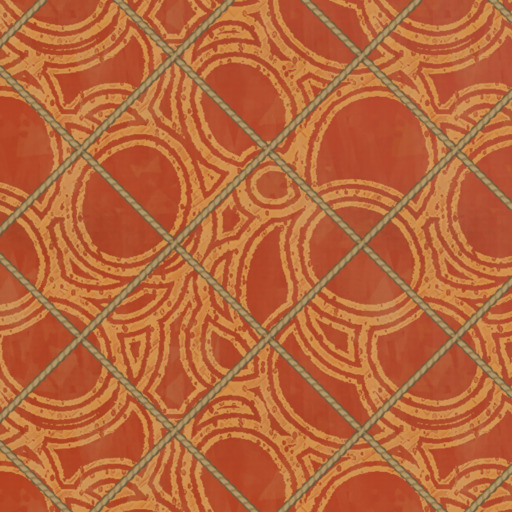} &
    \includegraphics[width=\q]{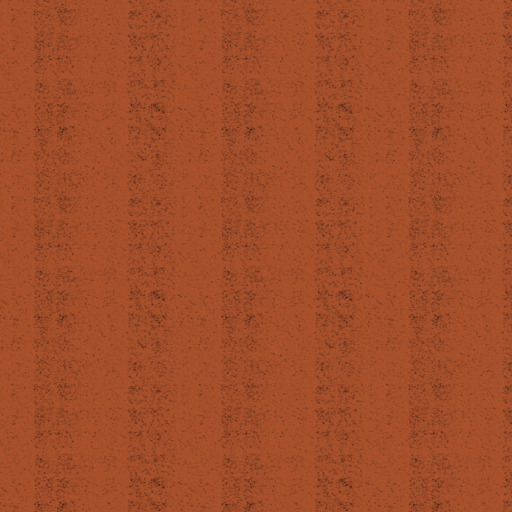} &
    \includegraphics[width=\q]{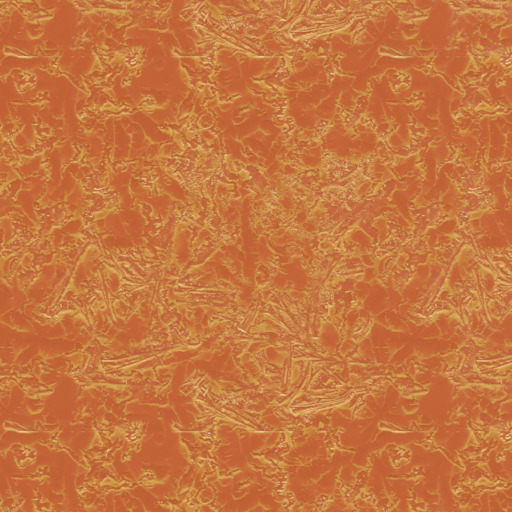} &
    \includegraphics[width=\q]{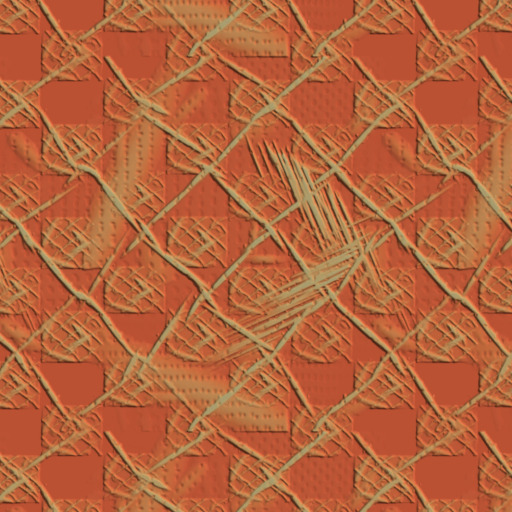}\\
    % \midrule
    \includegraphics[width=\q]                    {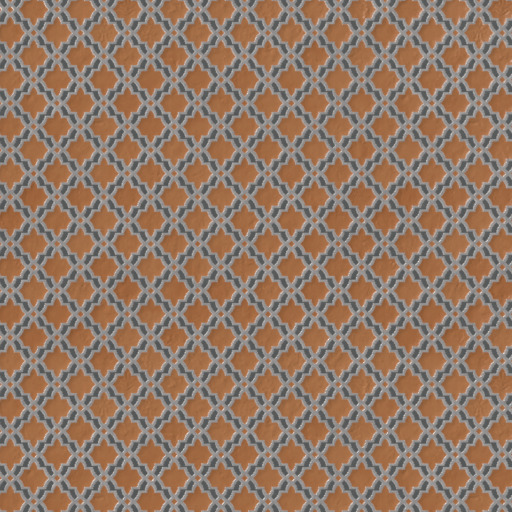} &
    \includegraphics[width=\q]{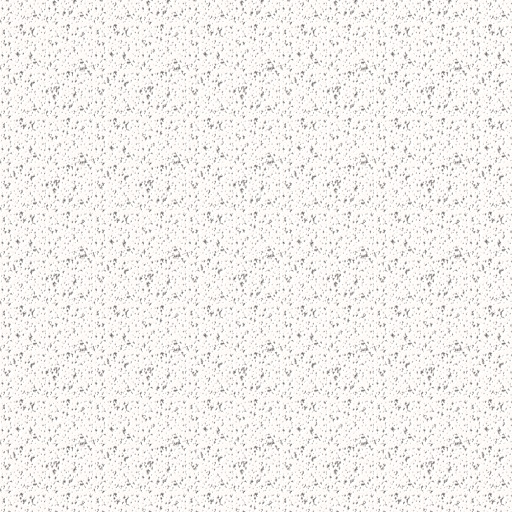} &
    \includegraphics[width=\q]{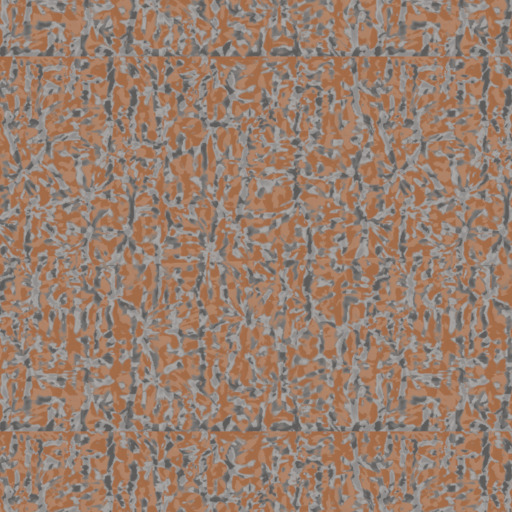} &
    \includegraphics[width=\q]{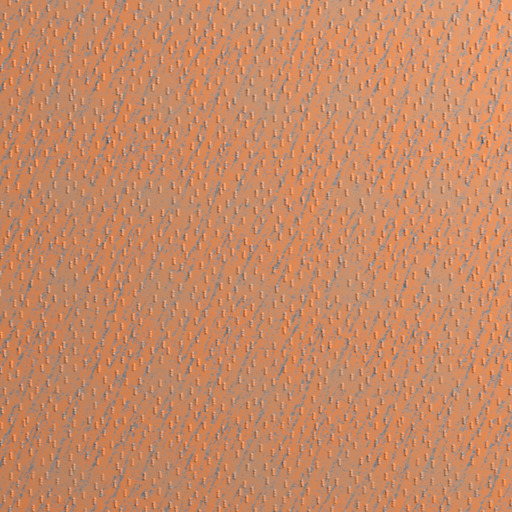}\\
    \bottomrule
  \end{tabular}
  \caption{Representative failure cases from the same challenging subset in
    \figfigref{fig:examples}{fig:more-examples}. All models struggle to
    reproduce the intricate patterns in these examples, though
    \projectname*[Graph] and \projectname*[Mixed] still outperform
    \vlmaterial*[SBS].}%
  \label{fig:failure-cases}
\end{figure}

\begin{figure}
  \pgfmathsetlength{\q}{\textwidth/4 -2\tabcolsep}
  \begin{tabular}{>{\centering\arraybackslash}m{\q} | >{\centering\arraybackslash}m{3\q}}
    \toprule
    \thead{Render} & \thead{Graph}\\
    \midrule
    \includegraphics[width=\q]{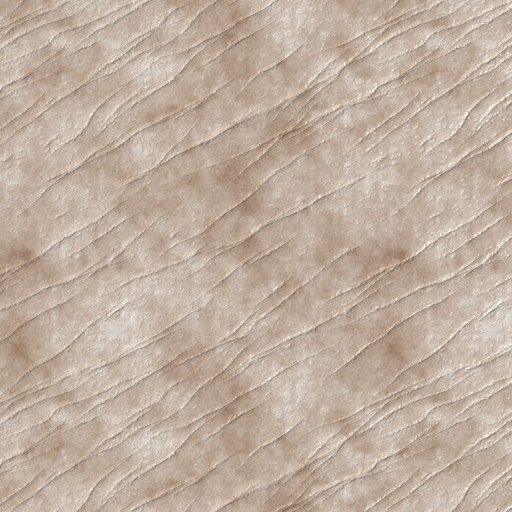} &
    \includegraphics[height=\q,width=3\q,keepaspectratio]{graphics/examples/unconditional/0/graph.pdf}\\
    % \midrule
    \includegraphics[width=\q]{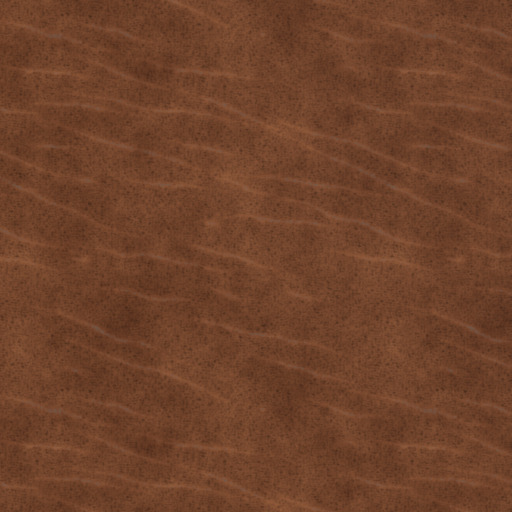} &
    \includegraphics[height=\q,width=3\q,keepaspectratio]{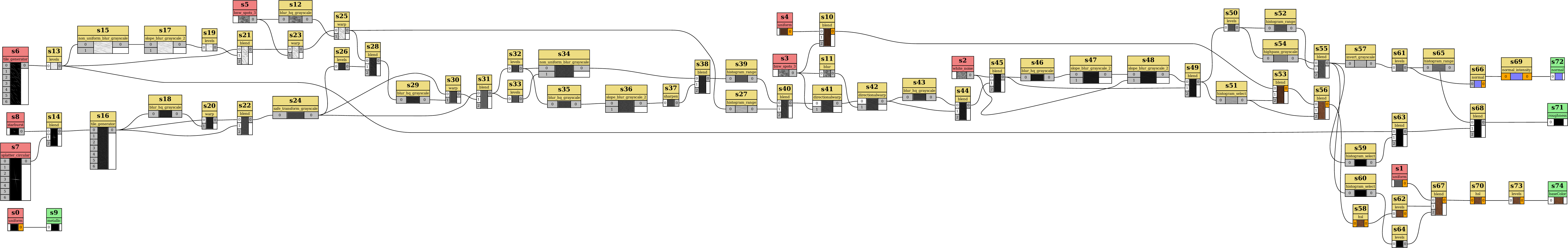}\\
    % \midrule
    \includegraphics[width=\q]{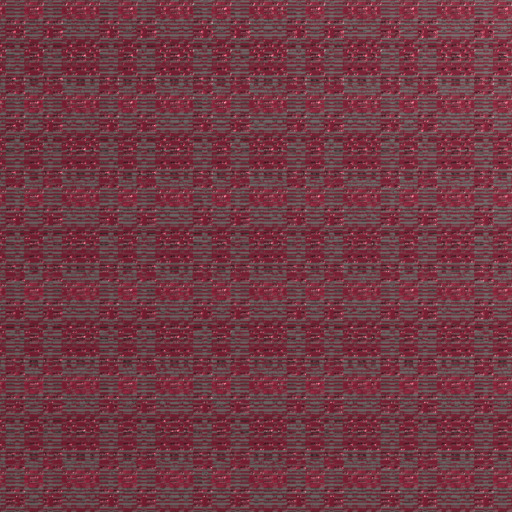} &
    \includegraphics[height=\q,width=3\q,keepaspectratio]{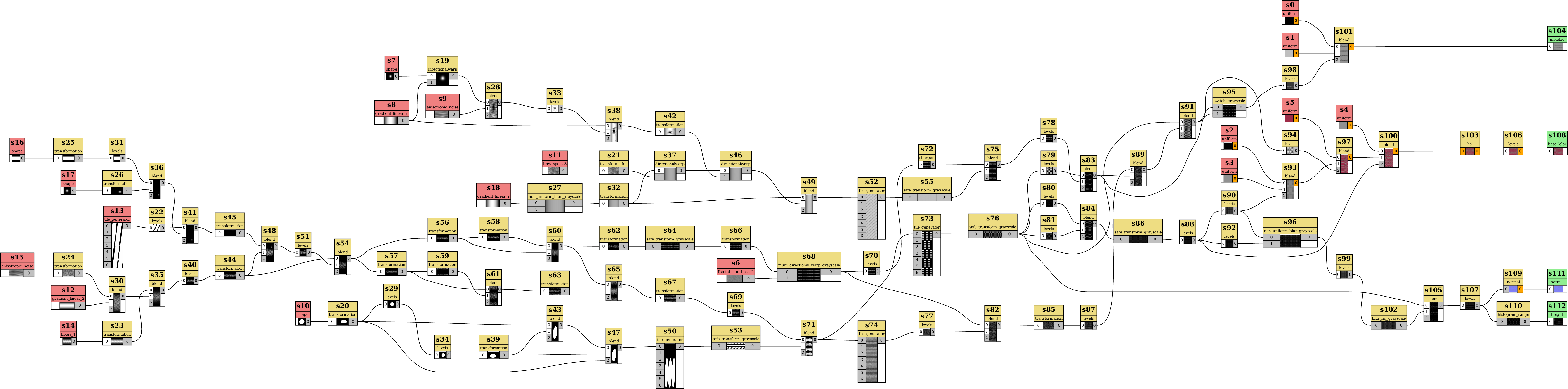}\\
    % \midrule
    \includegraphics[width=\q]{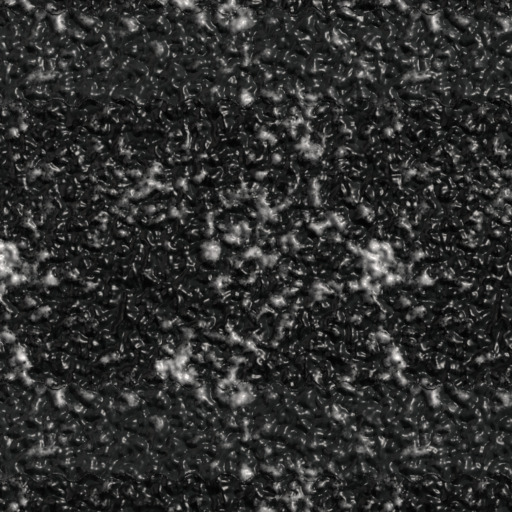} &
    \includegraphics[height=\q,width=3\q,keepaspectratio]{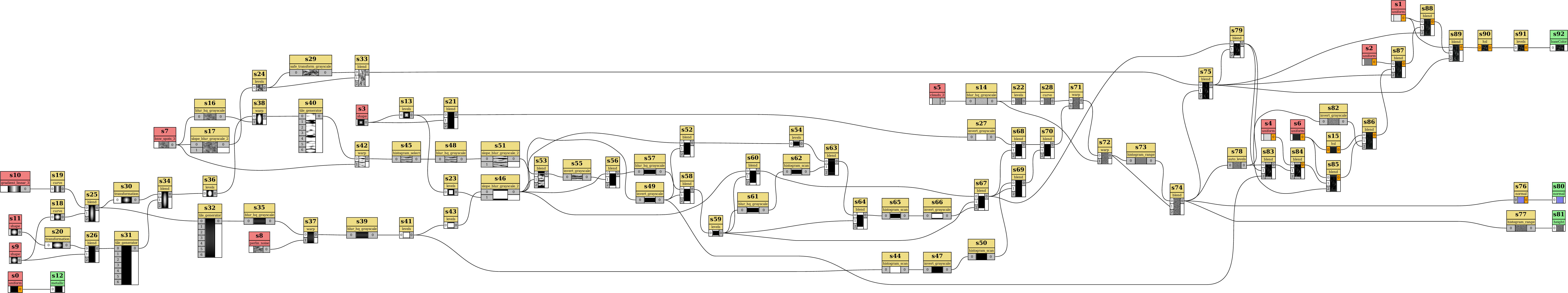}\\
    % \midrule
    \includegraphics[width=\q]{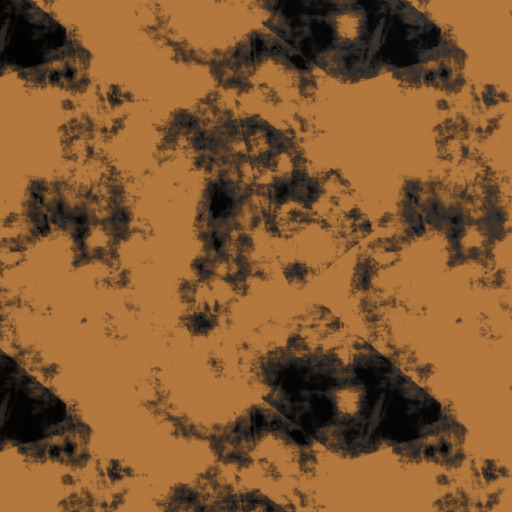} &
    \includegraphics[height=\q,width=3\q,keepaspectratio]{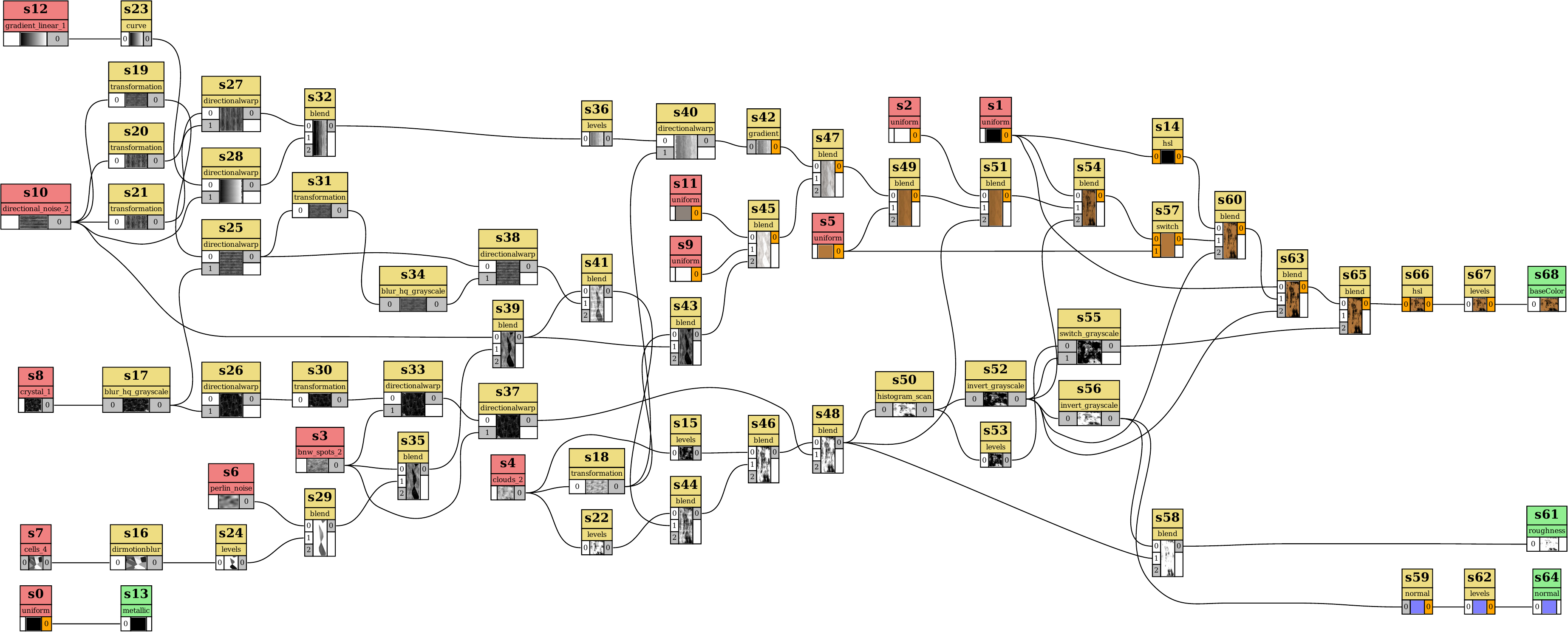}\\
    % \midrule
    \includegraphics[width=\q]{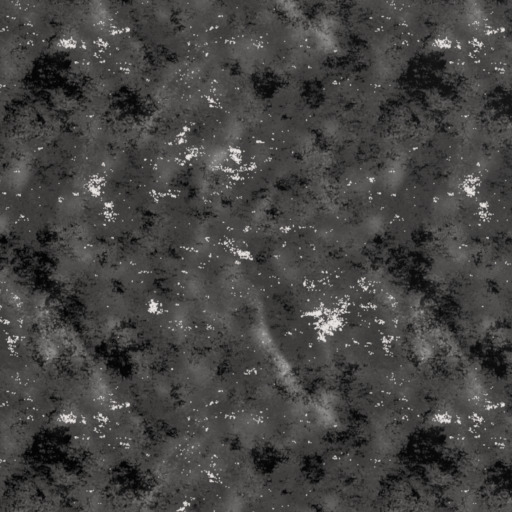} &
    \includegraphics[height=\q,width=3\q,keepaspectratio]{graphics/examples/unconditional/5/graph.pdf}\\
    \bottomrule
  \end{tabular}
  \caption{Example materials generated unconditionally by \projectname[Graph],
    shown alongside their corresponding procedural graphs.}%
  \label{fig:uncond-examples}
\end{figure}

In \figref{fig:more-examples} we provide additional qualitative
examples\@. \projectname*[Mixed] consistently surpasses \vlmaterial*[SBS],
while \projectname*[Graph] demonstrates the strongest results overall.
\figref{fig:failure-cases} complements
\figfigref{fig:examples}{fig:more-examples} by showcasing failure cases where
our models struggle to produce faithful outputs, though notably, the outputs
from \projectname*[Graph] and \projectname*[Mixed] still demonstrate superior
representation of the input compared to \vlmaterial[SBS]. Beyond these
conditional generation examples, \figref{fig:uncond-examples} presents
unconditional samples generated by \projectname[Graph], which exhibit high
visual quality with realistic material properties. Adjacent to these rendered
materials, we visualize their underlying material graphs in the same format
used as model input. In \figref{fig:code-examples}, we show a graph in
\sbs[Compact] representation to give an impression of the structure of our
format.

\begin{figure}
  \newtcbinputlisting{\sbslisting}[4][]{%%
    colframe=black,%
    size=small,%
    halign=center,%
    center title,%
    colback=white,%
    % tcbox width=auto limited,%
    % capture=hbox,%
    fontupper=\tcbfontsize{#3},%
    listing only,%
    minted language=yaml,%
    minted options={%
      style=bw,%
      breaklines=true,%
      breaksymbol={},%
      breakaftersymbolpre={},%
      breakafter={,},%
      breakindentnchars=4,%
      #2%
    },%
    listing file={#4},%
    #1%
  }%
  \begin{tcbraster}[raster columns=4,raster equal height=rows]
  \sbslisting{lastline=127}{.38}{graphics/examples/complete.yaml}%
  \sbslisting{firstline=128,lastline=254}{.38}{graphics/examples/complete.yaml}%
  \sbslisting{firstline=255,lastline=381}{.38}{graphics/examples/complete.yaml}%
  \sbslisting{firstline=382}{.38}{graphics/examples/complete.yaml}%
  \end{tcbraster}
  \caption{Complete example of a graph in \sbs[Compact] format. This listing
    shows the full representation of the material partially illustrated in
    \figref{fig:multimat-details}.}%
  \label{fig:code-examples}
\end{figure}
% This Figure is a transplied CompactSBS version of this Free asset from the platform: https://substance3d.adobe.com/assets/allassets/35db97cefcaeb6d388899792ee908ecd1d0867a3

\end{document}